\RequirePackage{tikz}
\pdfoutput=1
\documentclass{llncs}

\usepackage{mathtools}
\DeclarePairedDelimiter{\ceil}{\lceil}{\rceil}
\usepackage{tikz}
\usepackage{pgfplots}
\usepackage{xparse}
\usepackage{amsmath}
\usetikzlibrary{shapes.geometric, arrows}
\usepackage{multirow}
\usepackage{balance}
\usepackage{footmisc}
\usepackage[hyphens]{url}
\usepackage{graphicx}
\usepackage{array}
\usepackage{booktabs}

\usepackage{hyperref}
\usepackage{cleveref}
\usepackage{subcaption}
\usepackage{pifont}

\usepackage{geometry} 
\usepackage[margin=1cm]{caption}
\begin{document}

\title{Early Time-Series Classification Algorithms: An Empirical Comparison}

\author{Charilaos Akasiadis\inst{1} \and
        Evgenios Kladis\inst{1} \and
        Evangelos Michelioudakis\inst{1} \and
        Elias Alevizos\inst{1} \and
        Alexander Artikis\inst{1}}
\institute{Institute of Informatics and Telecommunications, NCSR `Demokritos', Agia Paraskevi, Greece \\\email{\{cakasiadis,eukla,vagmcs,alevizos.elias,a.artikis\}@iit.demokritos.gr}}

\maketitle

\begin{abstract}
Early Time-Series Classification (ETSC) is the task of predicting the class of incoming time-series by observing as few measurements as possible. Such methods can be employed to obtain classification forecasts in many time-critical applications. However, available techniques are not equally suitable for every problem, since differentiations in the data characteristics can impact algorithm performance in terms of earliness, accuracy, F1-score, and training time. We evaluate six existing ETSC algorithms on publicly available data, as well as on two newly introduced datasets originating from the life sciences and maritime domains. 
Our goal is to provide a framework for the evaluation and comparison of ETSC algorithms and to obtain intuition on how such approaches perform on real-life applications. The presented framework may also serve as a benchmark for new related techniques.
\end{abstract}


\section{Introduction} \label{sec:intro}

The evolution of computer systems and the integration of sensors and antennas on many physical objects facilitated the production and transmission of time-series data over the internet in high volumes~\cite{XuHCBHL14}. For instance, the integrated sensory and telecommunication devices on ships generate a constant stream of information, reporting trajectories in the form of time-series data \cite{FikiorisPAPP20}; the Internet of Things includes smart devices that constantly generate feeds of data related to localization, operational status, environmental measurements, etc. Such information can be exploited by machine learning techniques~\cite{EslingA12} to solve numerous everyday problems and improve a multitude of data-driven processes.

To that end, time-series classification methods have been applied on a plethora of use-cases, e.g., image \cite{RathM03} and sound \cite{HamooniM14} classification, quality assurance through spectrograms \cite{HolandKW}, power consumption analysis \cite{LinesBCA11}, and medical applications \cite{JambukiaDP}. Such methods train models using labeled, fully-observed time-series, in order to classify new, unlabelled ones, usually of equal length. 

Early Time-Series Classification (ETSC), on the other hand, extends standard classification, aiming at classifying time-series as soon as possible, i.e., before the full series of observations becomes available. Thus, the training instances consist of fully-observed time-series, while the testing data are incomplete time-series. In general, ETSC aims to maximize the trade-off between predictive accuracy and earliness. The objective is to find the earliest time-point at which a reliable prediction can be made, rendering  ETSC suitable for time-critical applications. 
For instance, in life sciences, simulation frameworks analyze how cellular structures respond to treatments, e.g., in the face of new experimental drugs~\cite{GiatrakosKDAGPA19}. Such simulations require vast amounts of computational resources, and produce gigabytes of data in every run. In the meantime, treatments which do not generate significant cell response could be detected at an early stage of the simulation and terminated before completion, thus freeing valuable computational resources.

In the maritime domain, popular naval routes around the globe require continuous monitoring, in order to avoid undesirable events, such as vessel collisions, illegal actions, etc~\cite{PitsikalisADRCJ19}. By utilizing available maritime time-series data, such events can be detected early in order to regulate naval traffic, or to take immediate action in case of suspicious behaviour. Similar examples stem from  the energy domain, where electricity consumption data can be analyzed to optimize energy supply schedules~\cite{TsoY}.

However, a prediction alone might not be useful by itself, since it should be incorporated in decision-making to become valuable~\cite{AtheyS}. Therefore, the earliness factor makes sense when there is enough time left to make effective use of a prediction. For instance, the ECG200\footnote{\url{http://www.timeseriesclassification.com/} \label{note}} dataset contains heartbeat data spanning $1$ second and the aim is to predict heart attacks; performing ECTS using, for example, $50\%$ of the time-series length, i.e., $0.5$ seconds of observations, most probably would not provide enough time for any proactive action. Moreover, many of the available time-series datasets are z-normalized. Z-normalized datasets are created using the mean and standard deviation calculated by all the time-points, those already obtained, and also those that are to be observed in the near future. This can be deemed unrealistic in ETSC applications~\cite{abs-2102-11487}, since values from the end of the time-series---required for the normalization step---would not be available at earlier time-points.  In summary, the datasets that are suitable for ETSC {\em (a)} should have a time horizon that would allow proactive decision-making, {\em (b)} should not be normalized, and {\em (c)} should have a temporal dimension (e.g., image shapes are not acceptable).

Although several ETSC methods have been proposed, there is a lack of experimental evaluation and comparison frameworks tailored to this domain. Existing reviews for time-series classification focus on comparing algorithms that do not generate early predictions. Representative reviews for standard time-series classification methods, as well as their empirical comparison can be found in ~\cite{AbandaML19},\cite{BagnallLBLK17},\cite{DhariyalNGI20},\cite{FawazFWIM19},\cite{RuizFLMB21}. Furthermore, ETSC methods are mostly evaluated and compared against only a few alternative algorithms. This is mainly performed using datasets from the UCR repository,\footnote{ \url{http://www.cs.ucr.edu/~eamonn/time_series_data_2018/}} which was originally created for evaluating standard classification approaches. As pointed in~\cite{abs-2102-11487}, most of the UCR datasets are z-normalized, meaning that values are altered by considering future time-points, which normally would not have been available in an online setting. This can introduce bias in the experimental results. In addition, many of these datasets do not include a temporal aspect (e.g., static shapes of objects), or they are only a few seconds long, thus not allowing any actual benefits from obtaining early predictions. A recent review of existing ETSC approaches is presented in~\cite{GuptaGBD20} featuring, however, a theoretical comparison and not an empirical one. 

To address this issue, we present an empirical comparison of ETSC algorithms on a curated set of meaningful datasets from real applications, providing insights about their strengths and weaknesses. We incorporate six algorithms i.e., {\em ECEC}~\cite{LvHLL19}, {\em ECONOMY-K}~\cite{DachraouiBC15}, {\em ECTS}~\cite{XingPY12}, {\em EDSC}~\cite{XingPYW11}, {\em MLSTM}~\cite{KarimMDH19}, {\em TEASER}~\cite{SchaferL20} into a publicly available and extensible Python framework.\footnote{\label{repo}\url{https://github.com/Eukla/ETS}} The algorithms are empirically evaluated in $12$ real-life datasets that fulfil the requirements of ETSC. Two of the datasets are new, originating from the drug treatment discovery and the maritime domains, while the remaining datasets are an appropriate subset of the well-known publicly available UEA \& UCR repository.\footref{note} Our empirical analysis shows that dataset size and high data variance can affect the performance of an ETSC algorithm. Overall, {\em TEASER}, {\em MLSTM}, and {\em ECEC} yielded better accuracy and earliness on most datasets, while {\em ECONOMY-K} was shown to be the fastest with respect to the time required for training. The presented experiments may be reproduced using our proposed framework.\footref{repo}

Our contributions may be summarized as follows:
\begin{itemize}
	\item We provide an open-source framework  for evaluating ETSC algorithms, which contains a wide spectrum of methods and datasets. Two of the included datasets are novel, from the fields of drug treatment for cancer and maritime situational awareness.
	\item We provide insights regarding the internal functionality of the examined ETSC algorithms by utilizing a simple running example.
	\item We empirically compare ETSC algorithms on a diverse set of appropriate datasets and outline the conditions that affect their performance.
\end{itemize}

The rest of the paper is organized as follows. In Section~\ref{sec:running_example} we present a running example that is used to explain algorithm functionality. 
In Section~\ref{sec:algorithms}, we describe the ECTS algorithms that we include in our framework. Then, Section~\ref{sec:datasets} presents an overview of the incorporated datasets. In Section~\ref{sec:experiments} we present our empirical evaluation, and in Section~\ref{sec:conclusion} we conclude.

\section{Running Example} \label{sec:running_example}

In order to aid the presentation of the algorithms throughout this paper, we use as a running example an excerpt of the dataset from the life sciences domain. This dataset comprises counts of tumor cells during the administration of specific drug treatments, resulting from large-scale model exploration experiments of the ways that the drug affects tumor cell growth~\cite{akasiadis2021parallel,Ponce-de-Leon2021.12.17.473136}. Each time-series represents one simulation outcome for a specific drug treatment configuration, characterized by the administration frequency, duration, and drug concentration. Each time-point in the resulting time-series corresponds to three different values, indicating the number of Alive, Necrotic and Apoptotic cells. Table~\ref{tb:running_example} shows a {\em prefix} of a randomly selected simulation. The length of a prefix can range from 1 up to the whole time-series length. In this example, we arbitrarily set the prefix length to 8.
{
\setlength\extrarowheight{2pt}
\begin{table}[h]
\centering
\begin{tabular}{|c|c|c|c|c|c|c|c|c|}
\hline
{\bf Time-point} & $t_0$ & $t_1$ & $t_2$ & $t_3$ & $t_4$ & $t_5$ & $t_6$ & $t_7$\\ 
\hline\hline
{\bf Alive cells}& 1137 & 1229 & 1213 & 1091 & 896 & 744 & 681 & 661\\ 
\hline
{\bf Necrotic cells} & 0 & 0 & 11 & 42 & 84 & 99 & 103 & 106\\ 
\hline
{\bf Apoptotic cells} & 0 & 1 & 17 & 118 & 282 & 432 & 509 & 549\\ 

\hline
\end{tabular}
\caption{Prefix of an actual simulation of cancer drug treatment.}
\label{tb:running_example}
\end{table}
}

Note that, in this case, the Alive tumor cells tend to decrease in number after the third time step, while the Necrotic cells are increasing, indicating that the drug is in effect. The Apoptotic cell count on the other hand, captures the natural cell death, regardless of the drug effect. Each of the time-series are labeled as \textit{interesting} or \textit{non-interesting}, based on whether the drug treatment has been effective or not, according to a classification rule defined by domain experts. The particular running example originates from a simulation classified as interesting, since the tumor shrinks (the number of alive cells decreases), as a result of applying a drug treatment.

\section{Overview of ETSC Algorithms} \label{sec:algorithms}
There is a number of different algorithms designed for ETSC that adopt different techniques for the analysis of the time-series. For example, the Mining Core Feature for Early Classification (MCFEC)~\cite{HeDPJQW15} utilizes clustering, the Distance Transformation based Early Classification (DTEC)~\cite{YaoLLZHGZ19} and the Early Classification framework for time-series based on class Discriminativeness and Reliability (ECDIRE)~\cite{MoriMKL17} rely on probabilistic classifiers, or the Multi-Domain Deep Neural Network (MDDNN)~\cite{HuangLT18} algorithm, which is based on neural networks.

Note though, that not every algorithm has an openly available implementation and, moreover, most require a number of domain-specific configurations. 
These facts make ETSC algorithms hard to apply in real-world settings, as the development- or domain expertise-related costs are increased.
For the purposes of our empirical comparison we focus on 6 available implementations that require the tuning of at most three configuration parameters.
In what follows, we proceed to describe the functionality of the algorithms that we incorporate, which are {\em EDSC}, {\em TEASER}, {\em ECEC}, {\em ECTS}, {\em ECONOMY-K}, and {\em MLSTM}.

\subsection{EDSC}

\begin{figure}
	\centering
	\includegraphics[width=0.46\textwidth]{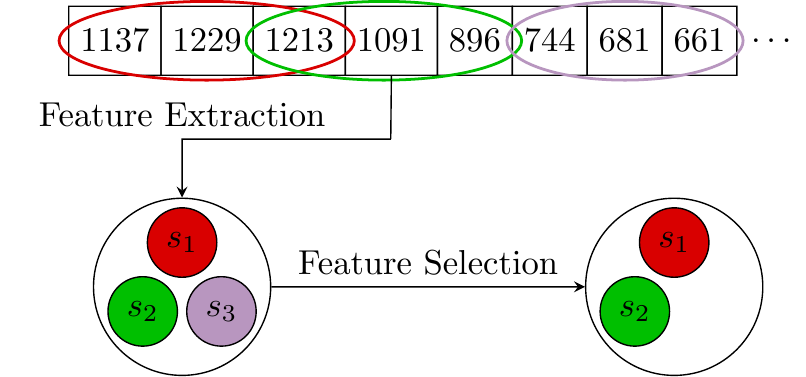}
	\caption{Illustration of {\em EDSC}.}
	\label{fig:EDSC}
\end{figure}

Early Distinctive Shapelet Classification ({\em EDSC})~\cite{XingPYW11} is one of the first methods proposed for ETSC. {\em EDSC} requires as input a user-defined range of subseries lengths that are used for shapelet extraction.
Shapelets are composed of subseries that are in some sense maximally representative of a class. Shapelets are defined as triplets of the form ($\mathit{subseries}$, $\mathit{threshold}$, $\mathit{class}$). 
The $\mathit{class}$ denotes the label of the time-series instance, as annotated in the training dataset, and the $\mathit{threshold}$ is the minimum distance that a time-series should have from the shapelet in order to be assigned to the same class. 
To compute the thresholds, {\em EDSC} finds all time-series that belong to different classes than that indicated by the shapelet's class, and measures their minimum distance from the shapelet's subseries. 
By utilizing the mean and the variance of these distances, as well as a user-specified parameter $k$ that fine-tunes the probability of a non-target time-series being matched to the target class, the Chebyshev's Inequality is applied to calculate the $\mathit{threshold}$. This ensures that, given a shapelet, each time-series excerpt with distance greater than the threshold indicates that the time-series belongs to a different class. 
During training, the algorithm isolates subseries from each input instance, and then calculates the thresholds to form the list of shapelets. 
Given this list, a utility function calculates a measure similar to the F$_1$-score for each shapelet, that represents its distinctive capability, i.e., how appropriate each shapelet is to be considered as a discriminator for a particular class. Ranked by their utility, the best shapelets are selected and grouped into a ``pool'' that constitutes the classification basis.

Consider the time-series of the Alive cells of the running example, as presented in Table~\ref{tb:running_example}, which is part of a dataset containing two classes, with labels $\mathit{zero}$ and $\mathit{one}$. Assuming that we are interested only in shapelets of length $3$, all possible subseries with that length are extracted. Figure~\ref{fig:EDSC} visualizes this example with three of the possible subseries. Let one subseries be $sb_{1} = \{1137,1229,1213\}$, originating from a time-series of class $\mathit{one}$. The minimum distance of a shapelet to a time-series is computed by aligning the shapelet against all subseries, and finding the minimum among their distances. Suppose that the minimum distance to $sb_{1}$ is stored in a list, the mean of which is $330$ and the variance $109$. Given $k = 3$, the threshold $\delta =  \mathit{max}\{ \mathit{mean} - k\cdot \mathit{var}, 0 \} = 3$ is calculated based on the Chebyshev's Inequality, and it indicates that time-series with distance less than $\delta$ from $sb_{1}$ belong to the same class. After this step, the shapelet $s_{1} = (sb_{1},\delta,\mathit{class})$ is created. The same procedure is repeated for the remaining subseries. Then, for each shapelet, a utility score is calculated, with the same formula as the F$_1$-score, but the weighted Recall~\cite{XingPYW11} is used instead of Recall. Then, the list is sorted and the top-$k$ shapelets that can accurately classify the whole training dataset are determined. 
Assume that for the three shapelets $s_{1}, s_{2}, s_{3}$ of Figure~\ref{fig:EDSC}, the list of utilities becomes $\{1.3, 3.67, 0.83\}$. The subseries of each shapelet are marked with ovals of different color. We first try to classify the whole training dataset using only $s_{2}$, since it has the highest utility. If we cannot correctly classify all the time-series in the dataset, then we add the second most ``valuable'' shapelet, $s_{1}$, to the set. Supposing that $s_{1}, s_{2}$ are informative enough, then we can claim that we have succeeded  in classifying the rest of the dataset, so the shapelet selection process is complete and the remaining shapelets are rejected, in this example $s_3$, as depicted in Figure~\ref{fig:EDSC}. When the minimum distance of a new, incoming time-series from a shapelet is less than $\delta$, then the shapelet's class is returned. This procedure is carried out for all possible prefixes of the incoming data, until a prediction is made.

{\em EDSC} supports only univariate time-series classification. Moreover, as the size of the dataset increases, so does the required time to extract and calculate shapelets. Thus, it is not expected to scale well for datasets with larger numbers of observations. For smaller datasets it trains quickly, with very low testing times, due to the simplicity of the incorporated classification procedure.
It should be noted that {\em EDSC} is one of the most widely used baselines for ETSC.

\subsection{TEASER}
\begin{figure}[b]
	\centering
	\includegraphics[width=0.46\textwidth]{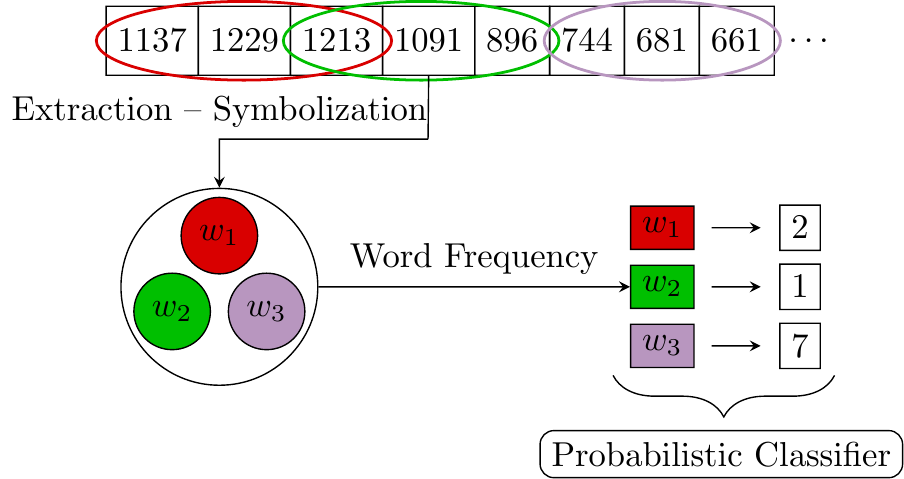}
	\caption{WEASEL procedure.}
	\label{fig:weasel}
\end{figure}

The Two-tier Early and Accurate Series classifiER method~\cite{SchaferL20} is based on the WEASEL classifier~\cite{SchaferL17}. WEASEL extracts subseries of a user-defined length and transforms them to ``words'' that are used to detect their frequency of appearance in the time-series. 
Figure~\ref{fig:weasel} illustrates this process. A time-series is passed to WEASEL, which subsequently extracts subseries (e.g. \{1137, 1229, 1213\}) and transforms them to symbols forming words ($w_1$, $w_2$, $\dots$). After this step, the frequencies of the words (e.g., assume \{2,1,7\}) are given to a logistic regression classifier.

 Notably, the training dataset is z-normalized and truncated into $N$ overlapping prefixes. 
The first prefix is of size equal to the length of the time-series divided by $N$, and the last one is the full time-series. For each prefix, a WEASEL-logistic regression pipeline is trained and then used to obtain probabilistic predictions.
 These predictions are then passed on to a One-Class SVM, uniquely trained for each prefix length. If the prediction is accepted by the One-Class SVM, i.e. it is marked as belonging to the class, then the last criterion to  generate the final output is the consistency of the particular decision for $v$ consecutive prefixes. The parameter $v$ is selected during the training phase by performing a grid search over the set of values $\{1,\dots, 5\}$. For each candidate value, the method tries to classify all the training time-series, and the one that leads to the highest harmonic mean of earliness and accuracy is finally selected. 
\begin{figure}[h]
	\centering
	\includegraphics[width=0.52\textwidth]{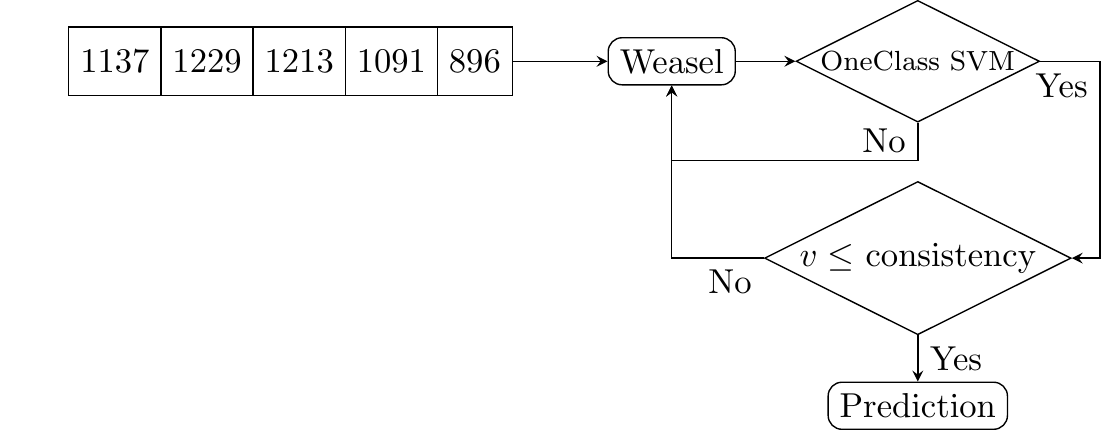}
	\caption{Illustration of {\em TEASER}.}
	\label{fig:TEASER}
\end{figure}
Figure~\ref{fig:TEASER} shows a schematic view of the procedure followed by {\em TEASER}. Similarly to the previous example, assume that the first examined prefix is of size $3$ and that the prediction made is accepted by the One-Class SVM. Assuming that $v=2$, the consistency check can only accept predictions that had been made for two consecutive prefixes. Thus, the current prediction will not be accepted, since in our case it was obtained by  using only one prefix. {\em TEASER} will wait for the next $2$ time-points. If {\em TEASER} does not manage to find an acceptable prediction by the time the final prefix arrives, then the prediction using the whole time-series is made without passing through the One-Class SVM or any other consistency check.

{\em TEASER} uses a pre-defined number of overlapping prefixes that reduces the number of possible subseries that need to be examined, thus boosting the method's performance. On the other hand, bad choices of $N$ can lead to suboptimal results. Also, it operates on univariate data and the z-normalization step that is inherently performed by {\em TEASER} regardless of dataset form, can be a major issue in the ETSC domain. Nevertheless, as we will see in Section~\ref{sec:experiments}, on average, {\em TEASER} achieves the best results with respect to earliness and the second best computation times, while attaining acceptable levels of accuracy, a result of the WEASEL symbolization process and the fast training of the logistic regressors and One-Class SVM.

\subsection{ECEC}

The Effective Confidence-based Early Classification algorithm ({\em ECEC})~\cite{LvHLL19} truncates the input into $N$ overlapping prefixes, starting from size equal to the length of the time-series divided by $N$ up to its full length, and trains $N$ classifiers (e.g. WEASEL) $H_{t}$. On this set of $N$ base classifiers, a cross validation is conducted, and the probabilistic predictions for each fold are obtained. Based on these probabilities, for each classifier $H_{t}$, {\em ECEC} measures the performance according to the probability of a true label being $y$, with the predicted label being $\hat{y}$, noted as $r_{H}(y\mid \hat{y})$. A core component of this algorithm is the calculation of the confidence threshold, which indicates the reliability of a prediction for each prefix size $k$:

\begin{equation*}
	C_{t}(H_{t}(X)) = 1 - \prod_{k=1}^t(1-r_{H_{k}}(y=H_{t}(X)\mid H_{k}(X))
\end{equation*}
where $X$ is the dataset, $t$ is the current time-point, and $H_{k}$ is the classifier at the $k$-th time-point. Based on the equation above, for each time-series and each prefix, {\em ECEC} calculates the confidence of the prediction made from the corresponding classifier $H_{t}$ during cross validation and stores it in a list. Then, the list of confidences is sorted, and the mean of adjacent values is saved as a threshold candidate ($\theta_{i}$) in a new list. For each $\theta_{i}$ and each time-series, the confidence of the classifier predictions at each time-point is compared to $\theta_{i}$. If a prediction is confident enough, {\em ECEC} stores it along with the time-point and the confidence value. Once all time-series for all prefixes are evaluated for $\theta_{i}$, {\em ECEC} checks the performance of the given threshold. The time-points and the predictions saved during the previous step are then used to calculate the accuracy and earliness for each threshold, according to the evaluation cost function $\mathit{CF}(\theta)$ value:
\begin{equation*}
	\mathit{CF}(\theta) = \alpha(1-\mathit{Accuracy}) + (1-\alpha)\mathit{Earliness}
\end{equation*}
$a$ is a parameter that allows users to tune the trade-off between accuracy and earliness. 
The $\theta_{i}$ that minimizes this cost, is marked as the global best threshold $\theta$.

\begin{figure}[t]
	\centering
	\includegraphics[width=0.52\textwidth]{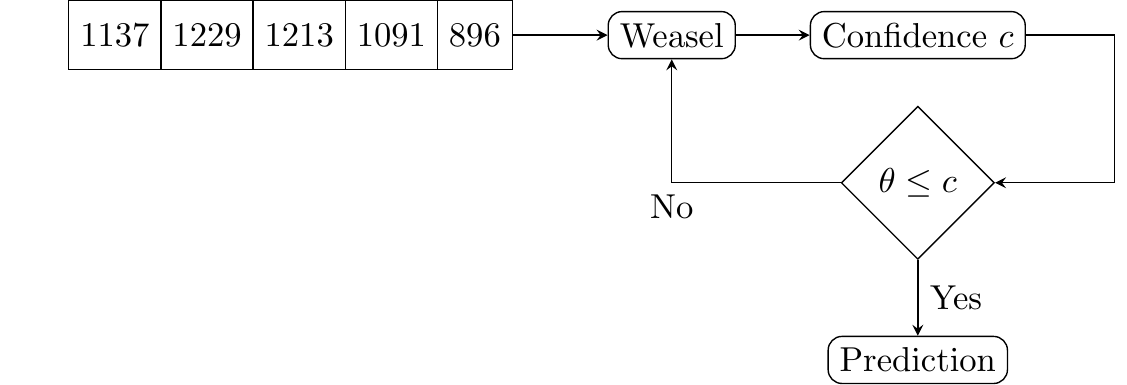}
	\caption{Illustration of {\em ECEC}.}
	\label{fig:ECEC}
\end{figure}

Figure~\ref{fig:ECEC} shows an example of the WEASEL used by {\em ECEC}. When a new input stream arrives, {\em ECEC} uses the minimum prefix size to make a prediction. Assuming that $N=2$ and the length of the time-series is 5, the minimum prefix during training is $\ceil{\frac{5}{2}}$, i.e. $\{1137,1229,1213\}$. Such prefixes are then passed to WEASEL, which in turn outputs the corresponding words and frequencies. Subsequently, the corresponding confidence of a prediction is calculated as $ c = C_{t}(H_{t}(X^{t}_{\mathit{test}}))$. Suppose that the confidence $c$ is 0.45 and the confidence threshold $\theta$ is $0.5$. In this case, the prediction is rejected and {\em ECEC} awaits for more data in order to form the prefix of bigger size. After the next $2$ time-points arrive, a new prediction is made and the confidence is recalculated and compared to $\theta$. If, $c \geq \theta$ the prediction is accepted, otherwise even more data is required. 

Similar to {\em TEASER}, {\em ECEC} extracts subseries, but their number is limited by a user-defined parameter, and {\em ECEC} supports only univariate time-series. As we show later in Section \ref{sec:experiments}, {\em ECEC} generates predictions with high predictive accuracy, though training times can be significantly impacted as the size of data increases.
Probabilistic approaches such as {\em ECEC} calculate threshold values by taking into account the performance of the algorithm on all classes for each prediction. As could be deemed natural, high similarity between time-series of different class and class imbalance can lead to very strict thresholds, and consequently to worse earliness performance.

\subsection{ECTS}

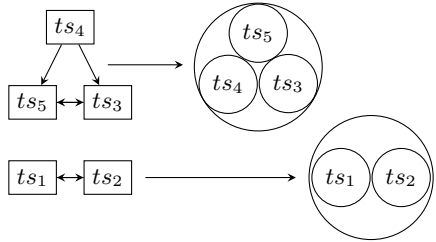
\begin{figure}[ht]
{\small 
	\centering
	\begin{tikzpicture}
	\node[draw] (T1) at (0,1) {$ts_1$};
	\node[draw] (T2) at (1,1) {$ts_2$};
	\node[draw] (T3) at (1,2) {$ts_3$};
	\node[draw] (T4) at (0.5,3) {$ts_4$};
	\node[draw] (T5) at (0,2) {$ts_5$};
	\draw[->,>=stealth] (T1) -- (T2);
	\draw[->,>=stealth] (T2) -- (T1);
	\draw[->,>=stealth] (T4) -- (T3);
	\draw[->,>=stealth] (T4) -- (T5);
	\draw[->,>=stealth] (T5) -- (T3);
	\draw[->,>=stealth] (T3) -- (T5);

	\draw[->,>=stealth]  (1.5,1) -- (3.5,1);

	\node at (4.5,1) [draw,circle,minimum size=1.65cm] {};
	\node at (4.1,1) [circle, minimum size=0.1cm, draw] {$ts_1$};
	\node at (4.9,1) [circle, minimum size=0.1cm, draw] {$ts_2$};

	\draw[->,>=stealth]  (1,2.5) -- (2,2.5);

	\node at (3,2.47) [draw,circle,minimum size=1.7cm] {};
	\node at (3.4,2.25) [circle, minimum size=0.1cm, draw] {$ts_3$};
	\node at (2.6,2.24) [circle, minimum size=0.1cm, draw] {$ts_4$};
	\node at (3,2.92) [circle, minimum size=0.1cm, draw] {$ts_5$};
	\end{tikzpicture}
		\caption{Reverse nearest neigbors (left) and clustering based on RNN and NN (right).}
	\label{fig:NN}
	}
\end{figure}

{\em ECTS}~\cite{XingPY12} is based on the $1$-Nearest Neighbor ($1$-NN) method. It stores all the nearest neighbor sets, for all time-series in the training set, and for each prefix length. Next, judging by the structure of these sets, it computes the Reverse Nearest Neighbors (RNN).
For example, in Figure~\ref{fig:NN}, a directed graph is visible on the left, with five time-series as nodes, which are prefixes of a given size. The in-degree of each node of the graph signifies how many RNNs it has. For instance, $ts_{5}$ has two inward edges; therefore two nodes consider it as their nearest neighbor, and thus $ts_{5}$ has RNN set = $\{ts_{4}$,$ts_{3}\}$. On the other hand, $ts_{4}$ has zero inward edges, so its RNN set is empty. For each possible prefix, this procedure is repeated to produce different RNN sets. The time-point from which the RNN set of a time-series remains the same until the end of the time-series, indicates that prefixes up to this time-point can be discerned from different class instances. This prefix length is called the Minimum Prediction Length (MPL). MPL signifies from which time-point onward a time-series can act as a classifier through the nearest neighbor search. For example, in Table~\ref{tb:NN}, the NN and RNN sets are presented for $5$ different time-series. The MPL (NN), of $ts_{2}$ means that $ts_{2}$ can act as a predictor for new time-series, using only the first $7$ time-points of  time-series.

In order to minimize the MPL and avoid needlessly late predictions, {\em ECTS} uses agglomerative hierarchical clustering. Time-series are merged based on the nearest neighbor distances to the clusters. Similarly, clusters are merged into larger ones, based on the Euclidean distances of their member items. The procedure continues as long as the new clusters contain same-label time-series, or until one cluster containing all the time-series remains. Each time a new cluster is formed, an MPL is assigned to it. The calculation of MPL is based on the RNN consistency, as well as on the $1$-NN consistency. The RNN sets of the cluster for each prefix, are calculated by applying relational division~\cite{ElmasriN15} on the union of the RNNs of the member time-series and the time-series of the cluster. Once the RNN set is constructed, {\em ECTS} finds the time-point from which then on the RNN set remains consistent. 

The $1$-NN consistency dictates that the nearest neighbor of each time-series in the cluster also belongs in that cluster for time-points up to the maximum length. The $1$-NN and RNN is calculated for each prefix. The time-point from which both $1$-NN and RNN sets are consistent up to the full time-series length, is the MPL of the cluster. During a cluster merging, time-series are accompanied with the smallest MPL among the cluster they belong to, and their own MPL. The result of the clustering phase is shown in Table~\ref{tb:NN}, where the MPL (Clustering) values are much lower for $ts_{2}$ and $ts_{3}$, making more accurate and earlier predictions. During the testing phase, for each prefix, new incoming time-series are matched to their nearest neighbor. If the observed length of the time-series up to the current time-point is larger than the MPL of its nearest neighbor, a prediction is returned.

Similarly to the previous algorithms, {\em ECTS} operates on univariate time-series. Due to the hierarchical clustering step, the method is sensitive to large, noisy and imbalanced datasets. Moreover, if time-series from different classes are similar, the clustering phase can be impacted, leading to higher MPLs. As we show later in our empirical evaluation, {\em ECTS} maintains a more stable performance across different dataset types with respect to earliness and accuracy.

\begin{table}[h]
	\begin{center}
	{
\setlength\extrarowheight{2pt}
		\begin{tabular}{|c|c|c|} \hline
		{\bf Time-Series} & {\bf MPL (NN)} & {\bf MPL (Clustering)} \\[1pt] \hline\hline
		$ts_1$ & 2 & 2 \\  \hline
		$ts_2$ & 7 & 3 \\  \hline
		$ts_3$ & 6 & 4 \\  \hline
		$ts_4$ & 4 & 4 \\  \hline
		$ts_5$ & 4 & 4 \\   \hline
		\end{tabular}
		}
	\end{center}
	\caption{Minimum prediction length calculation based on reverse nearest-neighbor and nearest-neighbor sets (left). Minimum prediction length optimization (right).}
	\label{tb:NN}
\end{table}

\subsection{ECONOMY-K}
Dachraoui et al~\cite{DachraouiBC15} introduce a decision function, which searches for the future time-point at which a reliable classification can be made. First, the full length training time-series are divided into $k$ clusters using K-Means. For each time-point a base classifier $h_t$ is trained (e.g., Naive Bayes or Multi-Layer Perceptron). For each cluster $k$ and time-point $t$, the classifiers $h_{t}$ are used to create a confusion matrix in order to compute the probability of a prediction being correct, $P_{t}(\hat{y}\mid y,k)$. 
\begin{figure}[ht]
{\small
	\centering
	\begin{tikzpicture}
	\usetikzlibrary{chains,fit,shapes,arrows,decorations.pathreplacing}
	\tikzstyle{process} = [draw,rounded corners,rectangle,text centered]
	\tikzstyle{decision} = [draw,diamond,text centered]
	\tikzstyle{arrow} = [->,>=stealth]
	\tikzstyle{line} = [-,>=stealth]
	
	\edef\sizetape{0.7cm}
	\tikzstyle{tmtape}=[draw,minimum size=\sizetape]
	\tikzstyle{tmhead}=[arrow box,draw,minimum size=.5cm,arrow box arrows={east:0.25cm, west:0.25cm}]
	\begin{scope}[start chain=1 going right,node distance=-0.15mm]
		\node [on chain=1,tmtape,draw=none] {};
		\node [on chain=1,tmtape] {1137};
		\node [on chain=1,tmtape] {1229};
		\node [on chain=1,tmtape] (time-m) {1213};
		\node [on chain=1,tmtape] {1091};
		\node [on chain=1,tmtape] (time-s) {896};
	\end{scope}
	\node at (1,-2) [draw,circle,minimum size=1.7cm] (one-cluster) {};
	\node at (0.55,-2) [circle, minimum size=0.08cm, draw]  {$ts_1$};
	\node at (1.35,-2) [circle, minimum size=0.08cm, draw]  {$ts_2$};
	
	\node at (3.5,-2) [draw,circle,minimum size=1.8cm] (two-cluster) {};
	\node at (3.9,-1.8) [circle, minimum size=0.07cm, draw] {$ts_3$};
	\node at (3.1,-1.8) [circle, minimum size=0.07cm, draw] {$ts_4$};
	\node at (3.5,-2.5) [circle, minimum size=0.07cm, draw] {$ts_5$};
	\draw[->,>=stealth]   (time-m.south) -- (one-cluster.north);	
	\draw[->,>=stealth]   (time-m.south) -- (two-cluster.north);	
	\end{tikzpicture}
	\hfil
	\begin{center}
	{
\setlength\extrarowheight{2pt}
		\begin{tabular}{|c|c|} \hline
			{\bf Cluster} & {\bf Membership Probability} \\ \hline\hline
			$cluster_1$ & 0.6 \\ \hline
			$cluster_2$ & 0.4 \\ \hline
		\end{tabular}
		}
	\end{center}
	\caption{Illustration of {\em ECONOMY-K}.}
	\label{fig:NM}
	}
\end{figure}
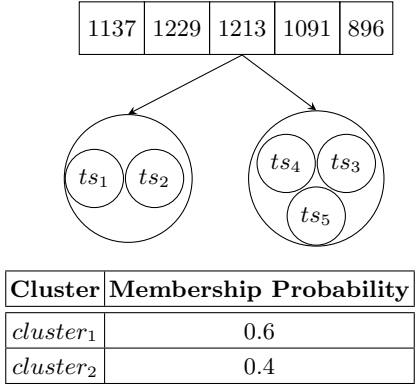
When incomplete time-series $ts_t$ enter the decision function, they are each appointed to a cluster membership probability: \[P(\mathit{cluster}_k \mid ts_t) = \frac{s_k}{\sum_{i=0}^{K} s_i}\] with $s_k$ being the following sigmoid functions:
\[s_k=\frac{1}{1+e^{-\lambda\Delta_k}}\] $\lambda$ is a constant and $\Delta_k$ is the difference between the average of all distances of the clusters to $ts_{t}$, and the distance of the cluster $k$ and $ts_t$. 
The goal of this method is to find the future time-point position $\tau$ out of the remaining ones i.e., in the next incoming time-points, where the prediction could be made with the best confidence. To that end, for each $\tau \in \{0,\dots,T-t\}$ the cost function is calculated as
\[
 f_\tau(ts_t) = \sum_{c_k}^{}P(c_k \mid ts_t)\sum_{y}^{}\sum_{\hat{y}}P_{t+\tau}(\hat{y} \mid y,c_{k})(\hat{y}\mid y) + C(t+\tau)
  \]
 $c_k$ is the $k$-th cluster and $C$ is an increasing cost function, relatively to the number of time-points. The time step $\tau$ with the least cost is the one when a prediction should be made. If the optimal $\tau$ is 0, then, the best time to make a prediction is at the time-point of the current measurement $t$. Therefore, the $h_t^k$ and $t$ are used to make a prediction, where $k$ is the cluster where the instance has the highest membership probability, and $t$ is the length of the examined time-series. For example, as presented in Figure \ref{fig:NM}, during the training step there are five time-series $ts_{1-5}$ that include a particular prefix, i.e. $\{1137,1229,1213,1091,896\}$, which is passed into {\em ECONOMY-K}. 
Then, using the K-Means step, two clusters are created, as well as the probabilities of the prefix belonging to either class. The future time-point position $\tau$ at which a prediction can be safely made, is either 0 or $1$, if the most appropriate time-point is the one expected to arrive next. Suppose that the cost function $f_\tau(ts_t)$ for each $\tau$, takes values $\{0.5, 1.2\}$. The minimum value is that of $f_0$, therefore the best time-point to make a prediction is the current one. Assume in the example of Figure \ref{fig:NM} that the highest membership probability belongs to $cluster_1$; therefore the classifier trained for the prefix's length and this cluster returns its prediction. If the smallest cost belonged to $\tau=1$, then the algorithm would need to wait for more data.

In general, clustering approaches are expected to be susceptible to noise, which can disrupt the creation of the correct clusters, and thus degrade performance.
However, the efficient implementations of K-Means and Naive Bayes allow {\em ECONOMY-K} to train very fast compared to the rest of the evaluated algorithms, even for very large datasets, as it becomes evident later in our empirical evaluation as well.

\subsection{MLSTM}

Neural network-based approaches have been applied to solve standard classification problems, i.e., to perform classification given the full-length time-series. However, they can be adapted to conduct ETSC as well, by supplying only the prefixes of the time-series in the input. Nevertheless, they cannot automatically detect the best time-point to give an early prediction, and thus, their earliness is fixed for each training. 
The optimization of the earliness parameter can lead to increasing computation times, taking also into account the size of the dataset.
Since the operation of neural networks can be considered well-known, here we omit the running example.
\begin{figure}[h]
	\centering
	\includegraphics[width=0.3\textwidth]{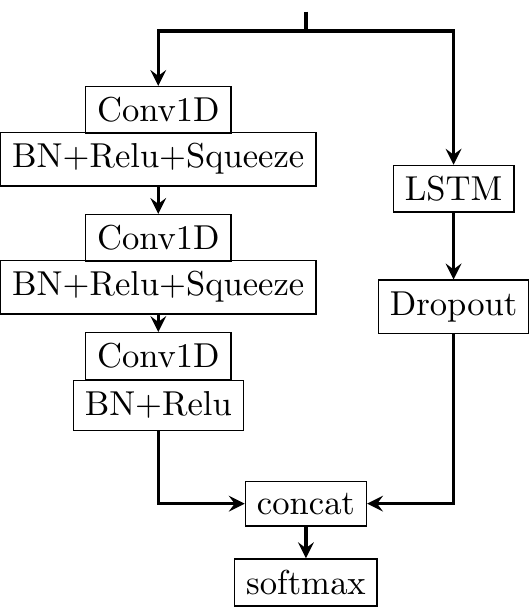}
	\caption{MLSTM architecture \cite{KarimMDH19}.}
	\label{fig:MLSTM}
\end{figure}
The Multivariate Long Short Term Memory Fully Convolutional Network ({\em MLSTM})~\cite{KarimMDH19} consists of two sub-models that are assigned the same input. 
The complete architecture is displayed in Figure~\ref{fig:MLSTM}. 
The first sub-model consists of three Convolutional Neural Network (CNN) layers. 
The use of CNNs is widely adopted in standard time-series classification~\cite{ZhaoLCLW17}, since it manages to extract important features from sequences of data. 
In this case, the outputs of each of the first two layers are batch normalized~\cite{IoffeS15} and are then passed on to an activation function, i.e., a Rectified Linear Unit (ReLU). 
In order to maximize the efficiency of the model for multivariate time-series, the activated output is also passed into a Squeeze-and-Excite block~\cite{HuSS18}, which consists of a global pooling layer and two dense layers that assign each variable of the time-series a unique weight. This has shown to increase the sensitivity of predictions~\cite{HuSS18}. 

The second sub-model comprises a masking layer and the output is passed on an attention-based LSTM. 
LSTM~\cite{KarimMDC18} is a popular Recurrent Neural Network model for time-series classification, because of its ability to `remember' inter time-series dependencies with minimal computational cost and high accuracy for time-series of length less than a thousand time points~\cite{BrownleeJ16}. 
Attention-based LSTMs are variations of the standard LSTMs with increased computational complexity, which, nevertheless, result to increased overall performance. 
The output of the two sub-models is concatenated and passed through a dense layer with as many neurons as the classes, and, via a softmax activation function, probabilistic predictions are generated.

{\em MLSTM} was introduced as a regular classification method, and was extended for ETSC by training it on time-series prefixes. 
Calculating the best LSTM cell number that maximizes the accuracy as well as the best prefix length is computationally demanding. 
Yet, {\em MLSTM} natively supports multivariate cases, in contrast to the algorithms presented so far, and, as shown in our empirical comparison, is a competitive method wrt. accuracy, F1-score, and earliness.

\begin{center}
{
\setlength\extrarowheight{3pt}
	\begin{table*}[h]
	\centering
		\begin{tabular}{|l|c|c|c|c|c|}
		\hline
			\multirow{ 2}{*}{{\bf Dataset Name}} & {\bf No. of Instances} & {\bf No. of Timepoints} & {\bf No. of} &  {\bf Class Imbalance }& {\bf Standard } \\
			{\bf } & {\bf  (Height)} & {\bf (Length) }  & {\bf Classes} & {\bf Ratio} & {\bf Deviation} \\

\hline
\hline
BasicMotions & 80 & 100  & 4 & 1 & 3.02 \\\hline	
Biological & 644 & 49 & 2 & 5.37 & 222.93 \\\hline
DodgerLoopDay & 158 & 288  & 7 & 1.25 & 12.77 \\\hline
DodgerLoopGame & 158 & 288  & 2 & 1.07 & 12.77 \\\hline
DodgerLoopWeekend & 158 & 288  & 2 & 2.43 & 12.77 \\\hline
HouseTwenty & 159 & $2{,}000$  & 2 & 1.27 & 779.19 \\\hline
LSST & $4{,}925$ & 36  & 14 & 111 & 167.26 \\\hline
Maritime & $5{,}249$ & 30  & 2 & 1.31 & 27.24 \\\hline
PickupGestureWiimoteZ & 100 & 361  & 10 & 1 & 0.2 \\\hline
PLAID & $1{,}074$ & $1{,}345$  & 11 & 6.73 & 3.41 \\\hline
PowerCons & 360 & 144  & 2 & 1 & 0.9 \\\hline
SharePriceIncrease & $1{,}931$ & 60  & 2 & 2.19 & 1.64 \\
\hline		\end{tabular}
		\caption{Dataset characteristics.}
		\label{tb:charact}
	\end{table*}
	}
\end{center}

\section{Datasets} \label{sec:datasets}

The datasets that we used in our evaluation consist of a subset of the publicly available UEA \& UCR Time Series Classification Repository~\cite{DauBKYZGRK19},
as well as of two new datasets that we introduce, from the life sciences and maritime domains. 
The selected datasets include both univariate and multivariate data. 
For the algorithms that cannot operate on multivariate cases, i.e., {\em ECTS}, {\em EDSC}, {\em TEASER}, and {\em ECEC}, different instances of the algorithm were trained for each variable, and a simple voting over the individual predictions was applied to obtain the final one.

\subsection{UEA \& UCR repository}

We selected $10$ out of $178$ UEA \& UCR datasets for our evaluation, according to the following criteria: {\em (a)} data should have a temporal dimension (e.g., image shapes are not acceptable), {\em (b)} data should not be normalized, and {\em (c)} the time horizon should be more than a few seconds. Note that several datasets from the UEA \& UCR repository have missing values or time-series of varying length, not allowing the application of most implemented ETSC methods. To address this, we filled in the missing values with the mean of the last value before the data gap and the first after it.

\subsection{Biological dataset: Cancer cell simulations}

This dataset originates from the life sciences domain, in particular drug discovery. To explore potentially helpful cancer treatments, researchers conduct large-scale model exploration with simulations of how tumor cells respond to drug administration~\cite{akasiadis2021parallel,Ponce-de-Leon2021.12.17.473136}. Each simulation is configured with a particular drug treatment configuration, and its course can be summarized by three time-evolving variables, i.e., the counts of three different tumor cell types for each time instant. Each experiment differs from the others based on a set of configurable parameters related to the treatment, i.e. the frequency of drug administration, its duration, and the drug concentration. These values remain fixed during each simulation. As explained in Sec.~\ref{sec:running_example}, each time-point of the resulting time-series corresponds to three different integer values, indicating the number of Alive, Necrotic and Apoptotic cells for each time step in the simulation experiment. The time-series are labeled as \textit{interesting} or \textit{non-interesting}, based on whether the drug treatment was found to be effective or not, i.e. managing to constrain tumor cell growth, according to a classification rule that was defined by domain experts. The dataset consists of $644$ time-series, each having $48$ time-points. The measurements were obtained by executing a parallel version of the PhysiBoSSv2.0 simulator.\footnote{\url{https://github.com/xarakas/spheroid-tnf-v2-emews}}

{
\setlength\extrarowheight{3pt}
\begin{table*}[h]
\centering
\begin{tabular}{|c|c|l|}
	\hline
	{\bf Group}& {\bf Specifications} & {\bf Datasets}\\  
	\hline
	\hline
	Wide& $\text{Length} > 1{,}300$ & HouseTwenty, PLAID  \\ 
	\hline
	Large&$\text{Height} > 1{,}000$& LSST, Maritime,  PLAID, SharePriceIncrease\\
	\hline
	Unstable& $\text{Std. Dev.} > 100$ & Biological, HouseTwenty, LSST   \\
	\hline
	\multirow{2}{*}{Imbalanced}& \multirow{2}{*}{Class Imbalance Ratio $>1$} & Biological, DodgerLoopDay, DodgerLoopGame, DodgerLoopWeekend,  \\
	&&   HouseTwenty, LSST, Maritime, PLAID, SharePriceIncrease \\
	\hline
	Multiclass& $\mbox{Number of Classes} > 2$& BasicMotions, DodgerLoopDay, LSST, PickupGestureWiimoteZ, PLAID   \\
	\hline
	\multirow{2}{*}{Common}& \multirow{2}{*}{None of the above}  &BasicMotions, DodgerLoopGame,  DodgerLoopWeekend,   \\ 
	&& PickupGestureWiimoteZ, PowerCons\\
	\hline
\end{tabular}
\caption{Categorization of datasets.}
\label{tb:data_categorization}
\end{table*}
}

In this dataset, classes are rather imbalanced. The \textit{interesting} time-series constitute the $20\%$ of the dataset, while the remaining $80\%$ accounts for \textit{non-interesting} cases. Also, many interesting and non-interesting instances tend to be very similar during the early stages of the simulation, until the drug treatment takes effect, which is usually after the first $30\%$ of the time-points of each experiment. Consequently, it is difficult to obtain accurate predictions earlier. For these reasons, this is a challenging benchmark for ETSC. 
 
\subsection{Maritime dataset: Vessel position signals}

The maritime dataset contains data from nine vessels that cruised around the port of Brest, France. This is a real dataset derived from~\cite{PatroumpasSCGPTSPT18,RayDCJ18}, and has been used for various computational tasks, such as complex event forecasting. Each measurement corresponds to a vector of values for longitude, latitude, speed, heading, and course over ground of a vessel at a given time-point. The time-series are fragmented to a specific length, and divided into two classes, based on whether the vessel did, or did not enter the port of Brest. Originally, the dataset was unlabelled and divided into nine time-series, one per vessel, each one having over $12{,}000$ time-points. In order to label them, we searched for the point in time that a vessel entered the port of interest and retrieved the previous $30$ time-points, thus creating positive examples of vessels that where actually entering the port. The rest of the observations that were not part of the positive examples were partitioned into $30$ time-point instances and assumed to belong to the negative class. In total, $5{,}249$ time-series instances were formed, each one having $30$ time-points, corresponding to $30$ minutes.
Apart from being multivariate (5 variables) and slightly imbalanced ($2{,}980$ negative and $2{,}269$ positive examples), the dataset includes the largest number of examples in our dataset list, making it a challenging application for ETSC.

\subsection{Categorization}

We categorize the selected datasets according to the characteristics that might impact algorithm performance. Table~\ref{tb:charact} presents the dataset characteristics. We measured the dataset size in terms of `length' and `height', where length refers to the maximum time-series horizon in the dataset (number of time-points per time-series) and height corresponds to the number of time-series instances. 
We also computed the standard deviation for each variable, as well as the class imbalance ratio, in order to detect unstable and imbalanced datasets.
This class imbalance ratio is calculated by dividing the number of instances of the most populated class with the number of instances of the least populated class. 
The number of classes was also considered, forming another category for the datasets that include more than two. 
The thresholds for height, length, and variance were set empirically, after examining the values for each dataset. 
This way, we end up with the six groups shown in Table~\ref{tb:data_categorization}. 
The first column of this table refers to the name of the dataset category, the second column shows the specifications for each category, and the third one lists all datasets belonging to each category. Note that the categories are not necessarily mutually exclusive, e.g. the HouseTwenty dataset belongs both to the `Wide' and the `Unstable' category.

\section{Empirical Comparison} 
\label{sec:experiments}
{
\setlength\extrarowheight{2pt}
	\begin{table}[h]
		\centering
		\begin{tabular}{|l|c|}
			\hline
			{\bf Methods} & {\bf Parameter values}\\
			\hline\hline
			{\em ECEC}& $N=20$, a$=0.8$\\
			\hline
			{\em MLSTM}& Attention-LSTM\\
			\hline
			{\em ECONOMY-K}& $k=\{1,2,3\}$, $\lambda$=100, $\mathit{cost} = 0.001$\\
			\hline
			{\em ECTS}& $\mathit{support}=0$\\
			\hline
			{\em EDSC}& CHE, $k=3$, $\mathit{minLen}=5$, $\mathit{maxLen}=L/2$\\
			\hline
			{\em TEASER}& $S:20$ for UCR, 10 for the Biological and Maritime \\
			\hline
		\end{tabular}
		\caption{Parameter values of ETSC algorithms.}
		\label{tb:parameters}
	\end{table}
}

\subsection{Experimental setting}
In our empirical evaluation we use accuracy and $F_{1}$-score to measure the quality of the predictions. Also, we employ the earliness score, defined as the length of the observed time-series at the time of the prediction, divided by the total length of the time-series. 
The harmonic mean between earliness and accuracy is also shown.  Note that since lower earliness values are better, in contrast to the accuracy where higher values are better, we invert the earliness values to $1 - \mbox{earliness}$ so that the result of the harmonic mean is reasonable.
Moreover, we present the training times for each algorithm. 
We compare the six algorithms incorporated in our framework, i.e, {\em ECEC}, {\em MLSTM}, {\em ECONOMY-K}, {\em ECTS}, {\em EDSC}, {\em TEASER}. 
Most implementations were readily available in Python, Java, or C++, except for {\em ECTS} that we had to implement ourselves. 
The experiments were performed on a computer operating in Linux, equipped with an Intel Xeon E5-2630 2.60GHz (24-cores) and 256 GB RAM. 

For the Biological and the UCR \& UEA datasets we performed a stratified 5-fold cross-validation over the classes, in order to facilitate comparison with published results. 
For the Maritime dataset, the folds were created so that each one contained the same number of instances from each vessel in order to address selection bias, as the trajectories of different vessels may differ significantly.

Since most algorithms don't support multivariate input, a voting method is applied, similar to the one employed in \cite{RuizFLMB21}. 
To that end, each classifier is trained and tested separately for each variable of the input time-series. Upon collecting each of the output predictions (one per variate), the most popular one among the voters is chosen, nevertheless assigned with the worst earliness among them. In the case of equal votes, we select the first one.

Regarding the configuration of each algorithm, {\em MLSTM} was tested on the $\{40\%, 50\%, 60\%\}$ of the time-series length for each training dataset, and the length yielding the best results according to the the harmonic mean of accuracy and earliness was selected. The number of LSTM cells was determined by using a grid search among $\{8, 64, 128\}$ for all experiments, choosing the one yielding the best score. For {\em TEASER}, the number of prefixes/classifiers $S$ was set to $10$ for the biological and maritime datasets whereas for the UCR \& UEA datasets it was set to $20$. For {\em ECEC}, the number of prefixes $N$ was set to $20$. These parameters for both {\em TEASER} and {\em ECEC} were chosen using manual, offline grid search. {\em TEASER-Z} applies z-normalization internally according to the original algorithm design, however, this might not be suitable for an online setting in the ETSC domain. Thus, we decided to also test a variant of this algorithm without the normalization step noted as {\em TEASER}. The $v$ parameter for {\em TEASER}'s consistency check is optimized for each dataset. Finally, for {\em ECONOMY-K} we experimented on $\{1, 2, 3\}$ clusters for each dataset. Table \ref{tb:parameters} summarizes all the parameter values used in our empirical comparison. Algorithms that did not produce results within 24 hours were terminated and were marked as requiring unreasonably long time for training.
The source code of our framework for {\em reproducing the presented experimental evaluation} is publicly available, accompanied by the respective datasets and algorithmic configurations.\footnote{\url{https://github.com/Eukla/ETS}}

\subsection{Experimental results}

\begin{figure}[h!]
\begin{subfigure}{\textwidth}
\centering
\includegraphics[width=\textwidth]{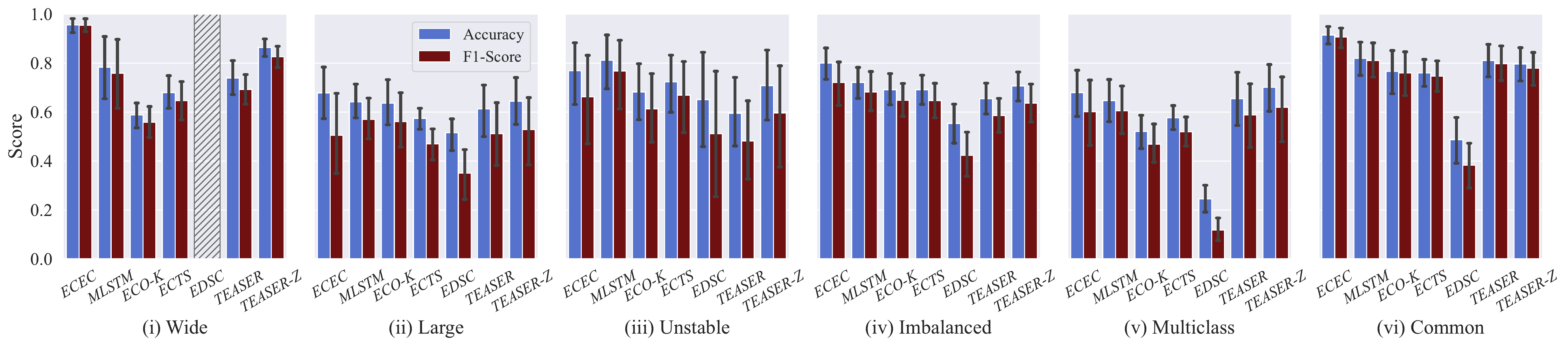}
\caption{Accuracy and F$_1$-Score.}
\label{fig:avg_acf1}
\end{subfigure}
\\
\begin{subfigure}{\textwidth}
\centering
\includegraphics[width=\textwidth]{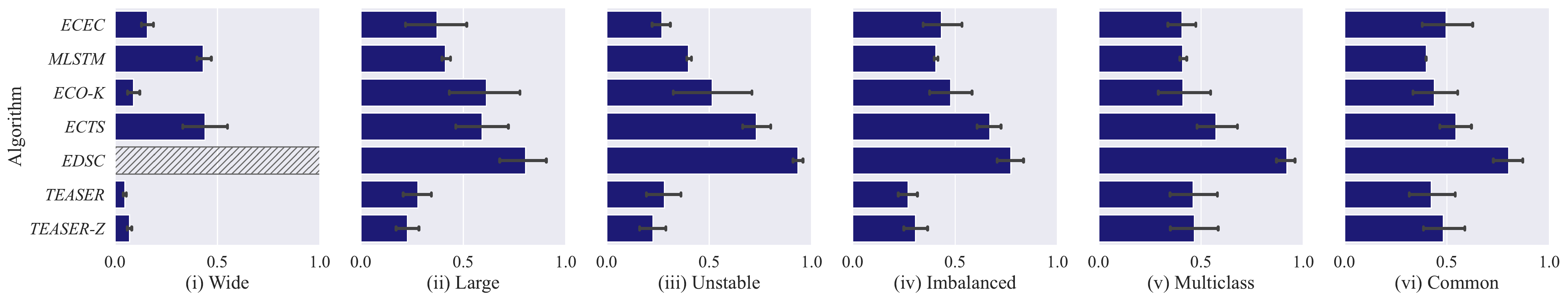}
\caption{Earliness (lower values are better).}
\label{fig:avg_earliness}
\end{subfigure}
\\
\begin{subfigure}{\textwidth}
\centering
\includegraphics[width=\textwidth]{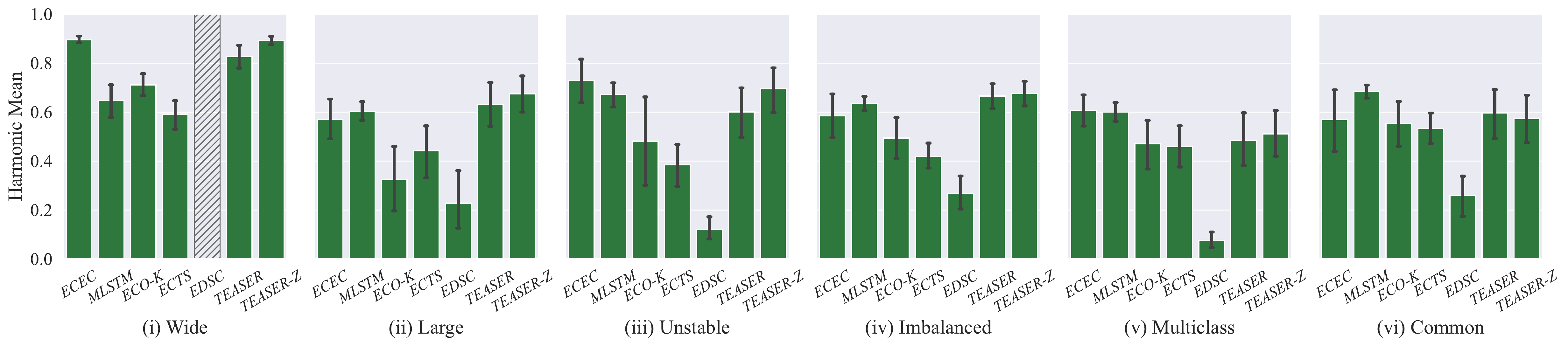}
\caption{Harmonic Mean between earliness and accuracy.}
\label{fig:avg_hm}
\end{subfigure}
\\
\begin{subfigure}{\textwidth}
\centering
\includegraphics[width=\textwidth]{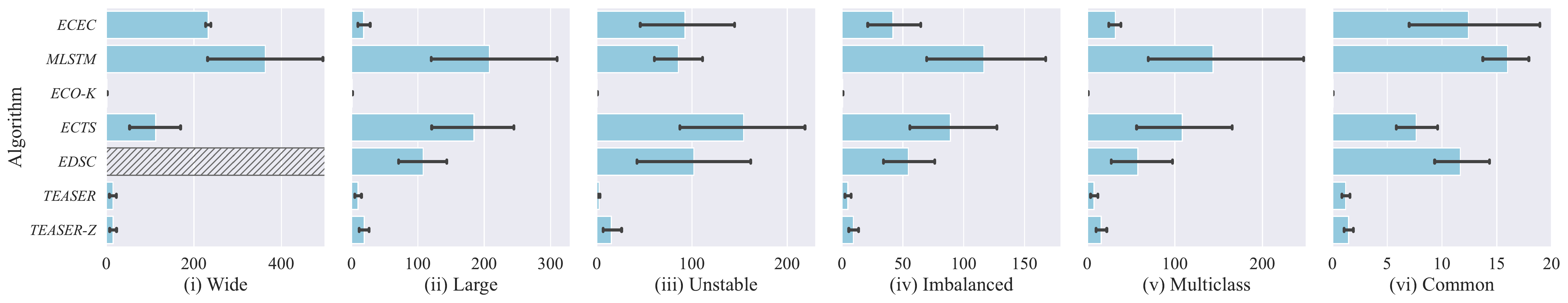}
\caption{Training times in minutes ($x$-axis scale varies).}
\label{fig:avg_time}
\end{subfigure}
\caption{Average scores per dataset category.}
\label{fig:catall}
\end{figure}

\begin{figure}[h]
\centering
	\begin{subfigure}{\textwidth}
		\centering
		\includegraphics[width=\textwidth]{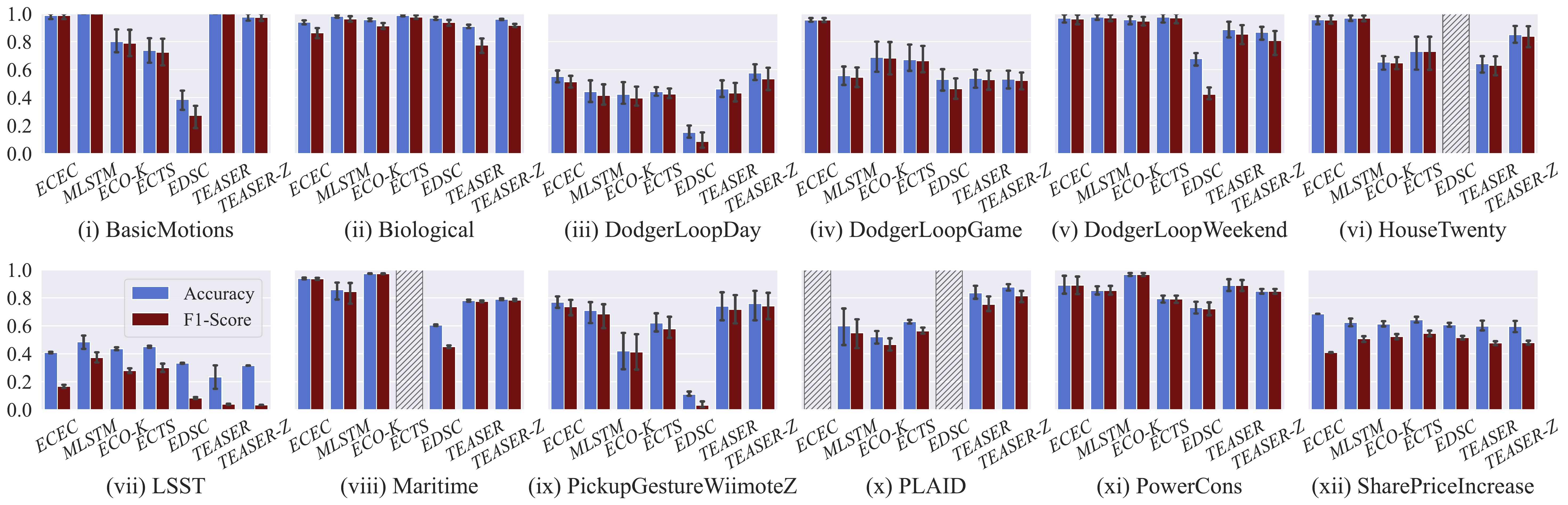}
		\caption{Accuracy and F$_1$-Score.}
		\label{fig:avg_AC_dat}
	\end{subfigure}
	\begin{subfigure}{\textwidth}
		\centering
		\includegraphics[width=\textwidth]{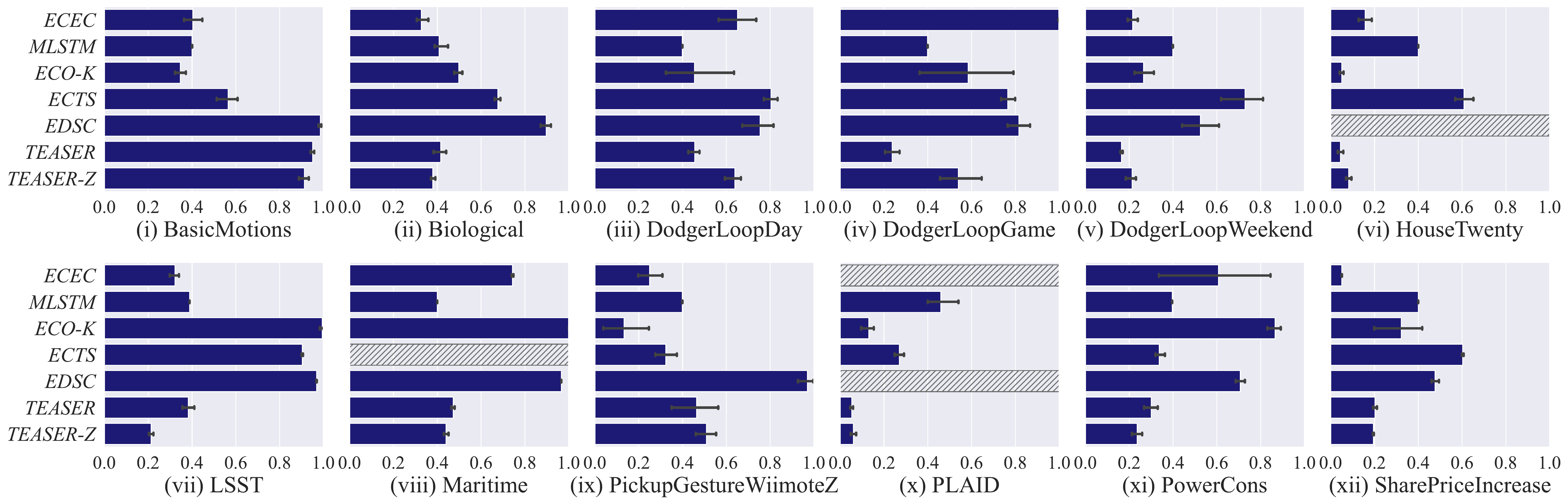}
		\caption{Earliness (lower values are better).}
		\label{fig:avg_Er_dat}
	\end{subfigure}
	\caption{Average predictive accuracy and earliness scores per dataset.}
	\label{fig:per_dat1}
\end{figure}

\begin{figure}[h]
	\begin{subfigure}{\textwidth}	
		\centering
		\includegraphics[width=\textwidth]{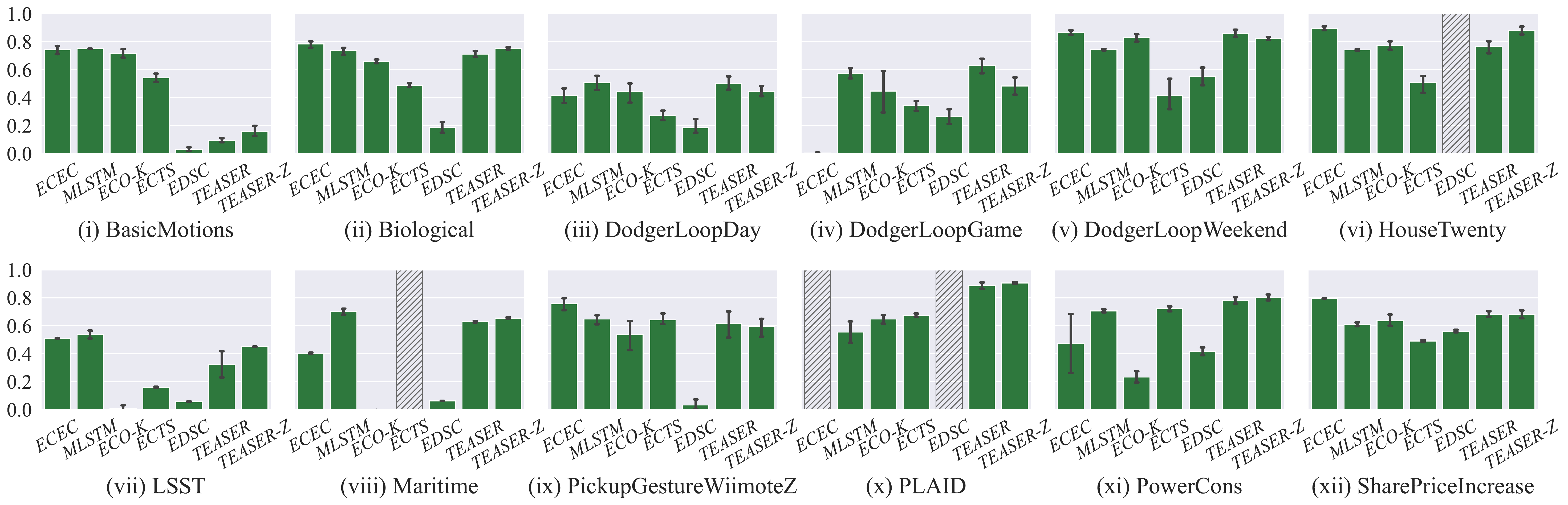}
		\caption{Harmonic Mean between earliness and accuracy.}
		\label{fig:avg_HM_dat}
	\end{subfigure}
	\begin{subfigure}{\textwidth}
		\centering
		\includegraphics[width=\textwidth]{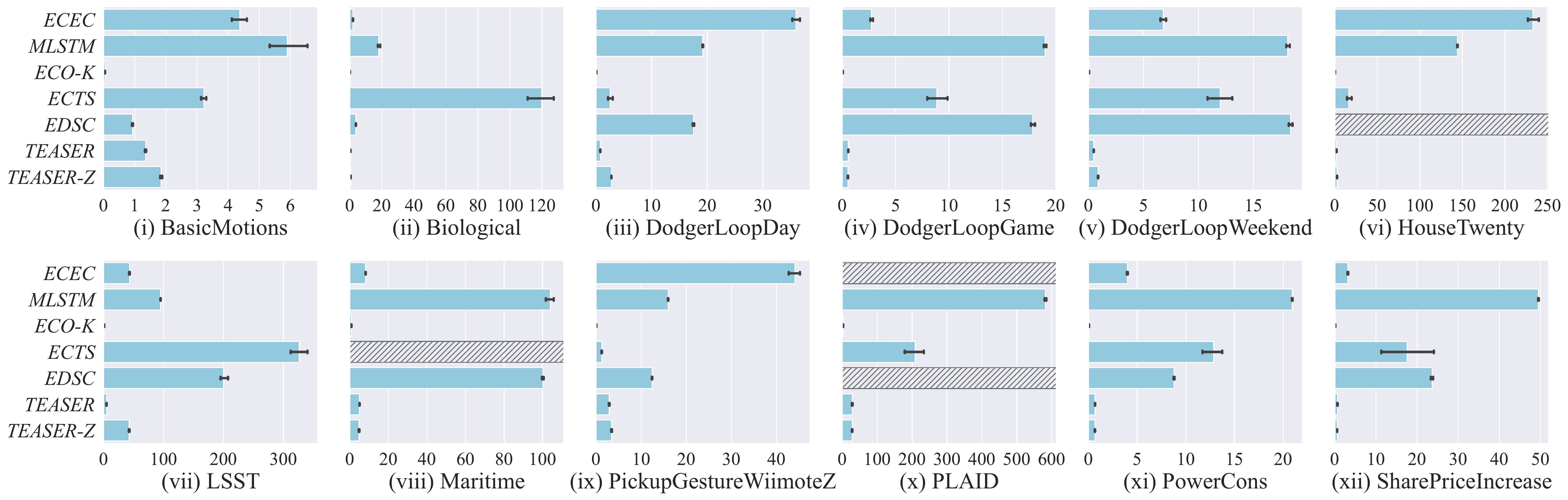}
		\caption{Training times ($x$-axis scale varies).}
		\label{fig:avg_TT_dat}
	\end{subfigure}
	\caption{Average harmonic mean scores and training time per dataset.}
	\label{fig:per_dat2}
\end{figure}

\begin{figure}[h]
\centering
	\begin{subfigure}{.24\textwidth}
		\centering
		\includegraphics[width=\textwidth]{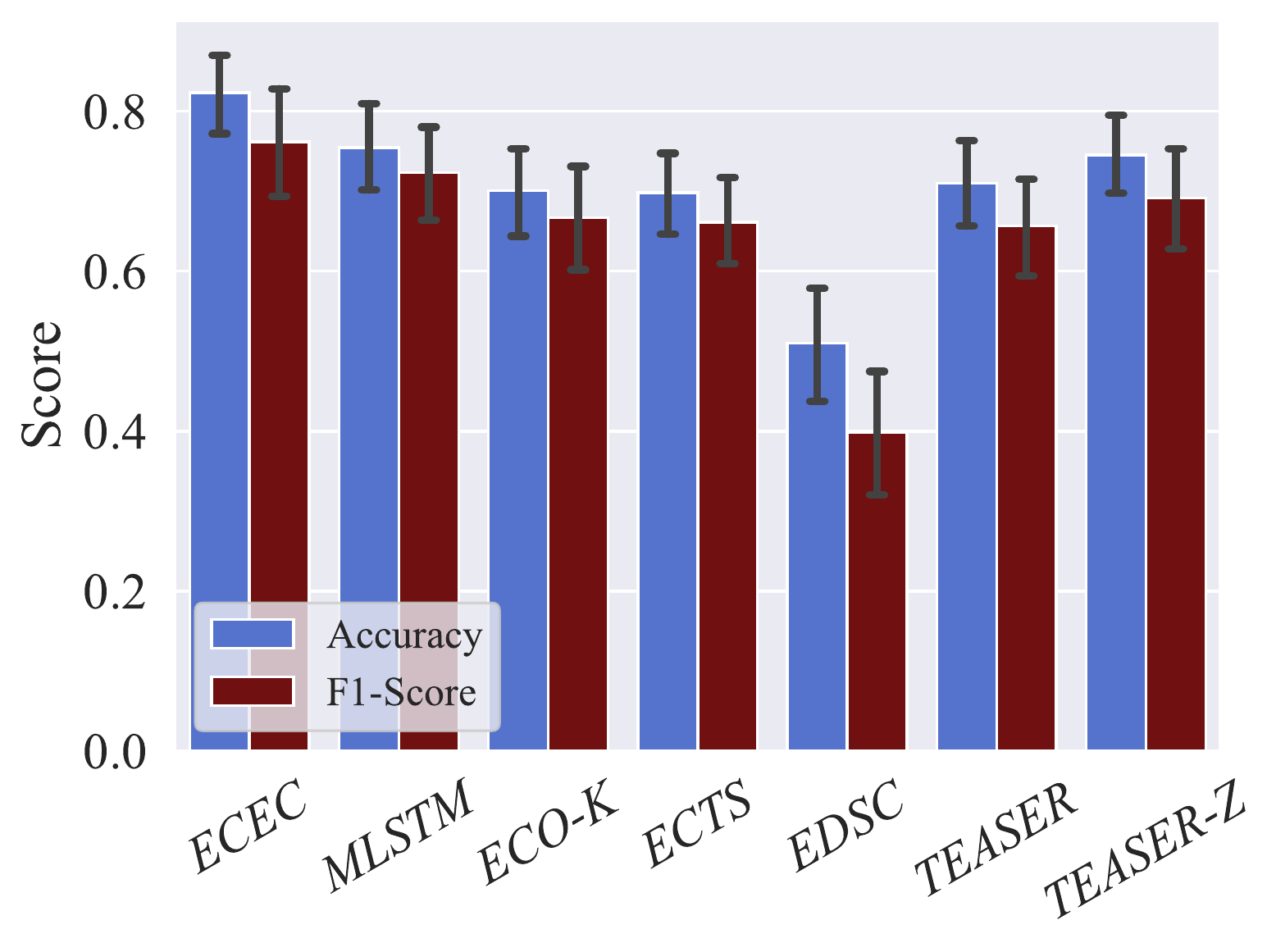}
		\caption{Accuracy and F$_1$-score.}
		\label{fig:accf1}
	\end{subfigure}
	\begin{subfigure}{.24\textwidth}
		\centering
		\includegraphics[width=\textwidth]{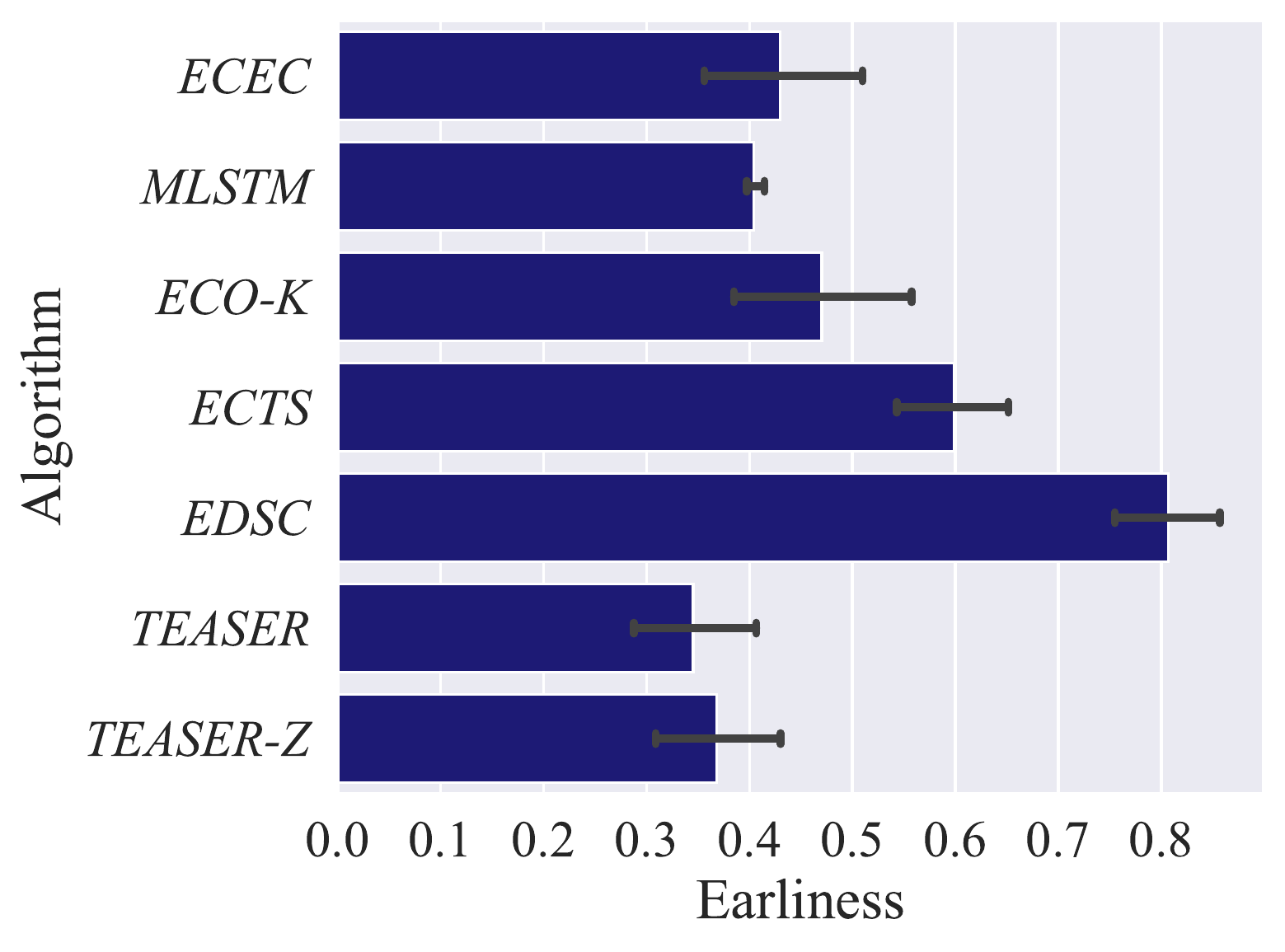}
		\caption{Earliness (lower values are better).}
		\label{fig:early}
	\end{subfigure}
	\begin{subfigure}{.24\textwidth}
		\centering
		\includegraphics[width=\textwidth]{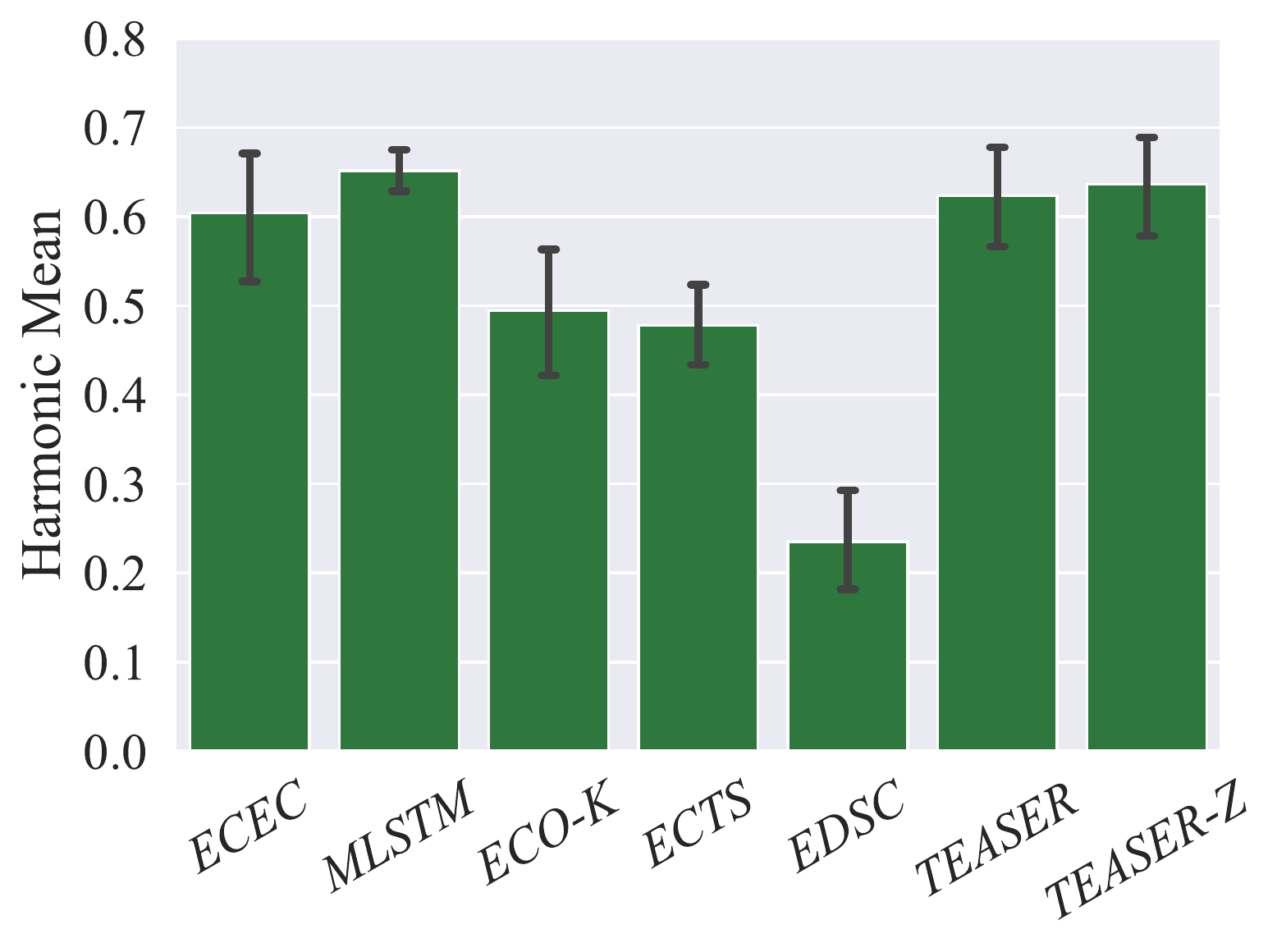}
		\caption{Harmonic Mean between earliness and accuracy.}
		\label{fig:harmmean}
	\end{subfigure}
	\begin{subfigure}{.24\textwidth}
		\includegraphics[width=\textwidth]{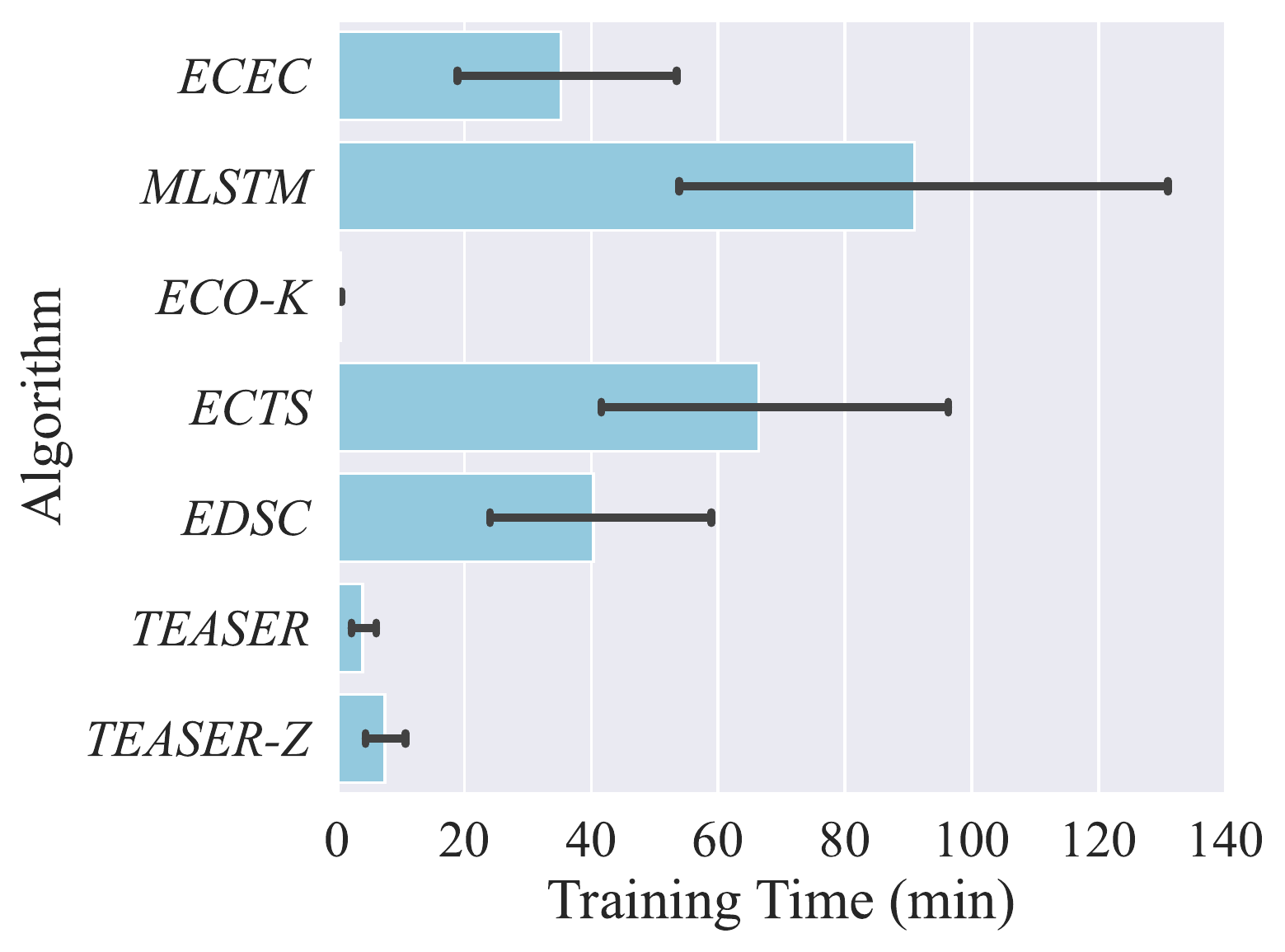}
		\caption{Training times.}
		\label{fig:time}
	\end{subfigure}
	\caption{Average scores across all datasets.}
	\label{fig:aggregate}
\end{figure}

Figure~\ref{fig:catall} presents the average performance for each algorithm on the six dataset categories.
Moreover, Figures~\ref{fig:per_dat1} and \ref{fig:per_dat2} display the performance of each algorithm per individual dataset, while Figure~\ref{fig:aggregate} presents the algorithms' performance across all datasets. In these figures, {\em ECONOMY-K} is abbreviated as {\em ECO-K} to save space. 
In the `Wide' datasets, i.e. HouseTwenty and PLAID, which both have more than $1{,}300$ time-points per instance, {\em ECEC} achieves the best accuracy and F$_1$-Score, followed by {\em TEASER-Z} and {\em MLSTM} (see Figure~\ref{fig:avg_acf1}(i)). Note though that {\em ECEC} did not manage to produce results for the PLAID dataset within reasonable time (see Figure~\ref{fig:avg_TT_dat}(x)). {\em TEASER}, {\em ECTS} and {\em ECONOMY-K} do not achieve an F$_1$-Score of more than 0.7, while {\em EDSC} did not manage to complete execution for neither of the `Wide' datasets (see Figures~\ref{fig:avg_AC_dat} and~\ref{fig:avg_TT_dat} (vi) and (x)). The large number of time-points allows {\em ECEC} to train robust confidence thresholds regardless of whether the underlying $k$ classifiers perform well or not.

In terms of earliness, in Figure~\ref{fig:avg_earliness}(i) we can see that the best score is achieved by {\em TEASER}, followed by {\em TEASER-Z}, {\em ECONOMY-K} and then {\em ECEC}. {\em MLSTM} and {\em ECTS} achieve worse earliness scores in this dataset category. Also, the One-Class SVM that the {\em TEASER} variants incorporate, helps them predict class labels with only a few time-points observed. 
The more time-points per instance a dataset has, the better earliness an algorithm can achieve, thus we observed overall increased performance for this dataset category by every algorithm.
Figure~\ref{fig:avg_hm}(i) shows the harmonic mean between accuracy and earliness, where {\em ECEC} and {\em TEASER-Z} have better overall performance, followed by {\em TEASER}, {\em ECONOMY-K}, {\em MLSTM} and {\em ECTS}.
The z-normalization step seems to favor {\em TEASER-Z} performance as compared to {\em TEASER}.
We also observe that z-normalization leads to slightly worse earliness, since larger value deviations among time-points, that might provide insight for classification, are now decreased.

Judging by the training times of Figure~\ref{fig:avg_time}(i), {\em ECONOMY-K} is the fastest and {\em TEASER} and {\em TEASER-Z} perform quite fast as well compared to the rest of the algorithms. 
{\em ECONOMY-K}'s Naive Bayes and K-Means outperform all algorithms in every dataset category, while the shapelet extraction steps of {\em EDSC} are significantly impacted by the number of time-points per dataset instance.

Next, for the `Large' dataset category, i.e. the datasets including more than 1000 instances, in Figure~\ref{fig:avg_acf1}(ii) we can observe a drop in the performance of most algorithms except for {\em ECONOMY-K}. {\em ECEC}, {\em MLSTM} and the {\em TEASER} variants also have a similar performance in terms of accuracy, but {\em MLSTM} and {\em ECONOMY-K} achieve slightly better F$_1$-Scores. 
The performance drop for this dataset category is due to the high similarity between instances that belong to different classes, impacting, e.g., the underlying WEASEL classifiers for {\em ECEC} and {\em TEASER}, and the hierarchical clustering of {\em ECTS}, which did not manage to produce results within reasonable time for the Maritime dataset (see Figure~\ref{fig:avg_TT_dat}(viii)).

Figure~\ref{fig:avg_earliness}(ii) shows the earliness scores, with {\em TEASER-Z} performing the best, followed by {\em TEASER}, {\em ECEC} and {\em MLSTM}. {\em ECONOMY-K}, {\em ECTS} and {\em EDSC} do not manage to provide predictions as early as the other algorithms for `Large' datasets. In general, {\em EDSC} equips a simple rule to decide the classification threshold, thus it uses most of the time-points in most datasets to make a prediction.
We can see that with respect to the earliness performance of the {\em TEASER} variants, the z-normalization step seems to help the One-Class SVM to be trained on a more dense and bounded space.
In terms of the harmonic mean between accuracy and earliness  {\em TEASER-Z} outperforms the other algorithms, followed by {\em TEASER}, {\em MLSTM}, and {\em ECEC}, while {\em ECTS}, {\em ECONOMY-K}, and {\em EDSC} perform worse (see Figure~\ref{fig:avg_hm}(ii)).

With respect to training times in Figure~\ref{fig:avg_time}(ii), {\em ECONOMY-K} is again the fastest, followed by the {\em TEASER} variants and {\em ECEC}. {\em EDSC} manages to produce results for this dataset category faster than {\em ECTS} and {\em MLSTM}.

The `Unstable' dataset category includes datasets where the standard deviation of each variable exceeds 100. 
Figure~\ref{fig:avg_acf1}(iii) shows that, in the `Unstable' category, {\em MLSTM} achieves the best accuracy and F$_1$-Score. {\em ECEC} follows closely, and {\em ECTS} ranks third. {\em ECONOMY-K} and {\em TEASER-Z} have a similar performance, however {\em TEASER-Z} illustrates increased variance in the F$_1$-Score. 
With respect to earliness, as shown in Figure~\ref{fig:avg_earliness}(iii),  {\em TEASER-Z} is better, followed by {\em ECEC} and {\em TEASER}, while {\em ECONOMY-K}, {\em ECTS} and {\em EDSC} perform worse. 
Concerning the harmonic mean between accuracy and earliness 
{\em ECEC} is the best option followed by {\em TEASER-Z} and {\em MLSTM} (see Figure~\ref{fig:avg_hm}(iii)). {\em TEASER} is slightly worse, but still better with a big difference than {\em ECONOMY-K}, {\em ECTS} and {\em EDSC}. 
We can see that the high variance in the dataset does not have a significant impact on the algorithms' performance.
In this dataset category, as shown in Figure~\ref{fig:avg_time}(iii), {\em ECTS} takes the longest to train, followed by {\em EDSC}, {\em ECEC} and {\em MLSTM}. 
On the other hand, {\em ECONOMY-K} is again the fastest, and {\em TEASER} is faster than {\em TEASER-Z} since it skips the z-normalization step.

The `Imbalanced' datasets are the ones that have a class imbalance ratio higher than 1, namely there are less instances for one or more particular classes than others. 
As we can see in Figure~\ref{fig:avg_acf1}(iv), {\em ECEC} illustrates the best accuracy and F$_1$-Score, followed by {\em MLSTM}. {\em ECONOMY-K} and {\em ECTS} perform similarly and the {\em TEASER} variants follow closely, with {\em TEASER-Z} performing slightly better than the version without z-normalization. {\em EDSC} ranks last, with larger difference between accuracy and F$_1$-Score than the rest of the algorithms.
With respect to earliness, Figure~\ref{fig:avg_earliness}(iv) shows that the best results are given by {\em TEASER}, followed by {\em TEASER-Z}, {\em MLSTM}, {\em ECEC} and {\em ECONOMY-K}. Again, the One-Class SVM of the {\em TEASER} variants helps to discern instances among classes earlier, with fewer observations.
On the other hand,  {\em ECTS} and {\em EDSC} do not manage to give a prediction by observing less than 50\% of the time-points of a problem instance.
What changes in this dataset category, is that {\em ECEC} illustrates improved accuracy and F$_1$-Score compared to the other algorithms, but this is not the case for the earliness scores too: imbalanced datasets with low variance lead to stricter thresholds, thus {\em ECEC} requires to observe more data in order to achieve improved accuracy and F$_1$-Score.
In Figure~\ref{fig:avg_hm}(iv) we observe that the low earliness results of the {\em TEASER} variants lead them ranking first in terms of the harmonic mean, followed by {\em MLSTM}, which has less varying results among the folds, and then {\em ECEC}, {\em ECONOMY-K}, {\em ECTS} and {\em EDSC}.
Moreover, {\em ECONOMY-K} is again the fastest, and {\em TEASER} and {\em TEASER-Z} train quite fast as well, as shown in Figure~\ref{fig:avg_time}(iv). The slowest algorithm is {\em MLSTM}, while {\em ECTS} and {\em EDSC} require on average more than 50 min. to train.
Finally, {\em ECEC} lies in the middle.

The fifth dataset category includes the `Multiclass' cases, i.e. datasets having more than two class labels as targets. 
Figure~\ref{fig:avg_acf1}(v) shows that {\em TEASER-Z} is able to achieve the most accurate predictions however, with not much difference between {\em TEASER}, {\em ECEC}, and {\em MLSTM}. 
In Figure~\ref{fig:avg_earliness}(v) we observe that {\em ECEC}, {\em MLSTM} and {\em ECONOMY-K} perform very similarly concerning earliness, followed by the two {\em TEASER} variants, while {\em ECTS} and {\em EDSC} require more than 60\% of the instance time-points to make a prediction. 
Figure~\ref{fig:avg_hm}(v) shows that {\em ECEC} and {\em MLSTM} reach the best harmonic mean scores between earliness and accuracy for the `Multiclass' datasets, indicating that these algorithms are appropriate for such cases. 
In Figure~\ref{fig:avg_time}(v) we can see that the algorithm ranking is more or less the same as that of the `Imbalanced' datasets, with {\em ECONOMY-K} being the fastest in training, and {\em MLSTM} the slowest, requiring more than 100 minutes in most folds.

In the `Common' datasets, that have no special characteristics that would classify them in the remaining dataset types,
 {\em ECEC} achieves the highest predictive accuracy (see in Figure~\ref{fig:avg_acf1}(vi)). 
{\em MLSTM} and {\em TEASER} follow, while {\em TEASER-Z}, {\em ECONOMY-K} and {\em ECTS} perform slightly worse. Again, {\em EDSC} ranks last in this category.
Figure~\ref{fig:avg_earliness}(vi) shows that {\em MLSTM} achieves the best earliness score, however without much difference from {\em TEASER} and {\em ECONOMY-K}. {\em TEASER-Z}, {\em ECEC} follow and {\em ECTS} and {\em EDSC} rank last, with {\em EDSC} being significantly worse.
For the `Common' datasets, {\em MLSTM} has the best harmonic mean score as shown in Figure~\ref{fig:avg_hm}(vi), while {\em TEASER}, {\em ECEC}, {\em TEASER-Z}, {\em ECONOMY-K} and {\em ECTS} illustrate a similar performance. The superiority of {\em MLSTM} in this case lies in the good earliness values of around 0.4, combined with the second best accuracy and F$_1$-Score that are close to 0.8, on average. The implicit design of LSTMs for sequence prediction enables {\em MLSTM} to perform quite well, but the lengthy process of training (see Figure~\ref{fig:avg_time}(vi)) combined with the grid search for the hyperparameters might render it inapplicable for some real world applications.
Figure~\ref{fig:avg_time}(vi) shows that the training time is reduced for all algorithms compared to the performance for the other dataset categories, with the fastest still being {\em ECONOMY-K}, followed by the {\em TEASER} variants, {\em ECTS}, {\em EDSC}, and {\em ECEC} and {\em MLSTM} ranking last.

For completeness, we also present the results for each dataset individually. Figure~\ref{fig:per_dat1} shows the accuracy, F$_1$-score, and earliness, while Figure~\ref{fig:per_dat2} shows the harmonic mean between earliness and accuracy, and the training times. 
As we can see in Figure~\ref{fig:avg_AC_dat}, for the SharePriceIncrease, LSST and DodgerLoopDay datasets, all algorithms have a hard time reaching accuracy and F$_1$-score of more than 70\%.  DodgerLoopDay is both `Multiclass' and `Imbalanced', SharePriceIncrease is `Large' and `Imbalanced', and LSST is `Large', `Unstable', `Imbalanced' and `Multiclass' (see Table~\ref{tb:data_categorization}). 
This indicates that the negative impact that dataset characteristics induce, can be of additive nature. 
In Figure~\ref{fig:avg_Er_dat} we can see that apart from {\em MLSTM}, all other algorithms illustrate varying earliness performance, due to the different mechanisms they use to determine the right time to generate a prediction, and to the difficulties posed by the dataset characteristics. For example, in the PLAID and HouseTwenty datasets the {\em TEASER} variants perform very good, as opposed to, e.g., BasicMotions and PickupGestureWiimoteZ.
With respect to the harmonic mean between earliness and accuracy, Figure~\ref{fig:avg_HM_dat} shows that in 11 out of the 12 dataset the {\em TEASER} variants are very competitive compared to the other algorithms; for BasicMotions, which is the only exception, the bad harmonic mean scores are due to the bad earliness scores, since in terms of accuracy and F$_1$-score their performance is close to optimal. 
This happens because the time-series belonging to different classes are quite similar between one another, not allowing {\em TEASER}'s consistency check to be satisfied early enough.
Furthermore, we observe that the Maritime dataset is more challenging than the Biological one, since it requires significantly longer time for training and the algorithm scores of harmonic mean between earliness and accuracy are considerably lower.

Figure~\ref{fig:accf1} presents the average scores across all datasets. With respect to predictive accuracy, {\em ECEC} performs the best on average, followed by {\em MLSTM} and {\em TEASER-Z}.
{\em ECONOMY-K}, {\em ECTS}, and {\em TEASER} exhibit quite similar performance, with {\em EDSC} ranking last. 
Note that the $\alpha$ configuration that we used for {\em ECEC} favors accuracy over earliness and
the results of Figure~\ref{fig:early} show that, on average, {\em TEASER} and {\em TEASER-Z} achieve the best earliness values followed by {\em MLSTM} and {\em ECEC}. The good earliness results for the {\em TEASER} variants are mainly due to the effectiveness of One-Class SVM. {\em MLSTM} manages to select the earliest configuration of 0.4, as a result of the predictive capability of LSTMs.
 {\em ECONOMY-K}'s performance is close, and {\em ECTS} and {\em EDSC} rank last, with {\em EDSC} not managing to give results by observing any less than 75\% of the time-series points. 
Figure~\ref{fig:harmmean} displays the average harmonic mean values;  {\em MLSTM} achieves the best balance between accuracy and earliness, while the  {\em TEASER} variants and {\em ECEC} follow closely. 
{\em ECONOMY-K} achieves an average of 0.5 among all datasets, and {\em ECTS} follows closely, while {\em EDSC} ranks last. 
Algorithms based on clustering techniques, i.e. {\em ECTS} and {\em ECONOMY-K}, have a hard time converging to both early and accurate models. 
Figure~\ref{fig:time} displays the average training times. This figure shows that {\em MLSTM} takes the longest to train, while the fastest is {\em ECONOMY-K}, followed by the {\em TEASER} algorithms. 

\section{Summary and Further Work} \label{sec:conclusion}

We empirically evaluated six state-of-the-art early-time series classification algorithms on benchmark datasets.
Moreover, we introduced and employed two datasets from the domains of cancer cell simulations and maritime situational awareness. 
We summarized the functionality of each algorithm using a simple running example in order to illustrate their operation. 
We divided the incorporated datasets to six categories and
observed that the size of the dataset, the presence of multiple classes, as well as class imbalance, induce fluctuations in performance.

Overall, the {\em TEASER} variants are shown to have the best scores concerning the tradeoff between earliness and accuracy, with the lowest training times. 
{\em MLSTM} achieves, on average, the best harmonic mean score between earliness and accuracy, however with increased training times.
{\em ECEC} is quite close with respect to the harmonic mean, and requires less time for training than {\em MLSTM}.
On the other hand, {\em ECONOMY-K} completes training very quickly, but it does not perform as well as the previous three with respect to the other performance metrics.
{\em ECTS} achieves harmonic mean scores close to {\em ECONOMY-K}'s, but is the second slowest in training after {\em MLSTM}.
{\em EDSC} does not perform as well as the other algorithms with respect to predictive accuracy and earliness, and it trains slower than {\em TEASER}, {\em ECEC}, and {\em ECONOMY-K}, but still faster than {\em MLSTM} and {\em ECTS}.

When training on datasets with larger numbers of instances, the performance of all algorithms is compromised, while datasets with more time-points per instance do not affect performance as much, as the harmonic mean scores increase for most algorithms, apart from {\em MLSTM} and {\em ECONOMY-K}. {\em ECEC} is quite competitive with respect to predictive accuracy and earliness for the `Wide', `Unstable', and `Multiclass' datasets, but still slower than {\em ECONOMY-K} and {\em TEASER} in training.
Regarding the z-normalization step of {\em TEASER-Z} compared to the variant without it, we detected better performance of {\em TEASER} for the `Common' datasets, but for the other categories {\em TEASER-Z} yelled slightly better results.

The repository that we developed for the presented empirical comparison includes implementations of all algorithms, as well as all datasets, and is publicly available,\footref{repo} allowing for experiment reproducibility. Moreover, the repository may be extended by new implementations and datasets, thus facilitating further research.  

In the future, new algorithms will be added to this framework, such as RelClass~\cite{ParrishAGH13} and DTEC~\cite{YaoLLZHGZ19}. 
Moreover, we plan to incorporate to our framework hyper parameter tuning techniques~\cite{ottervanger2021multietsc} for optimizing the configuration of early time-series classification algorithms.

\section*{Acknowledgment}

We would like to thank the reviewers of the SIMPLIFY-2021 EDBT workshop for their helpful comments. This work has received funding from the EU Horizon 2020 RIA program INFORE under grant agreement No 825070.

\bibliographystyle{splncs04}
\bibliography{references}

\end{document}